\documentclass{article}

\usepackage{microtype}
\usepackage{graphicx}
\usepackage{booktabs} % for professional tables

\usepackage{hyperref}

\usepackage{hyperref}       % hyperlinks
\usepackage{url}            % simple URL typesetting
\usepackage{amsfonts}       % blackboard math symbols
\usepackage{nicefrac}       % compact symbols for 1/2, etc.

\usepackage{subcaption}
\usepackage{mathtools}
\usepackage{multicol}
\usepackage{xspace}
\usepackage{dsfont}

\usepackage{soul}

\usepackage{lipsum}

\usepackage{amsmath,amsfonts,bm}

\def\eqref#1{equation~\ref{#1}}
\def\1{\bm{1}}

\DeclareMathAlphabet{\mathsfit}{\encodingdefault}{\sfdefault}{m}{sl}
\SetMathAlphabet{\mathsfit}{bold}{\encodingdefault}{\sfdefault}{bx}{n}

\DeclareMathOperator*{\argmin}{arg\,min}

\graphicspath{{figures/}}

\DeclarePairedDelimiter\abs{\lvert}{\rvert}%
\DeclarePairedDelimiter\norm{\lVert}{\rVert}%
\makeatletter
\let\oldabs\abs
\def\abs{\@ifstar{\oldabs}{\oldabs*}}
\let\oldnorm\norm
\def\norm{\@ifstar{\oldnorm}{\oldnorm*}}
\makeatother

\newcommand{\z}{\mathbf{z}}

\newcommand{\x}{\mathbf{x}}
\newcommand{\J}{\mathbf{J}}
\newcommand{\G}{\mathbf{G}}
\DeclareMathOperator{\MF}{MF}

\newcommand{\methodreg}{VHP-FMVAE\xspace}

\usepackage{xcolor}

\usepackage[accepted]{icml2020}

\icmltitlerunning{Learning Flat Latent Manifolds with VAEs}

\begin{document}

\twocolumn[
\icmltitle{Learning Flat Latent Manifolds with VAEs}

% It is OKAY to include author information, even for blind
% submissions: the style file will automatically remove it for you
% unless you've provided the [accepted] option to the icml2020
% package.

% List of affiliations: The first argument should be a (short)
% identifier you will use later to specify author affiliations
% Academic affiliations should list Department, University, City, Region, Country
% Industry affiliations should list Company, City, Region, Country

% You can specify symbols, otherwise they are numbered in order.
% Ideally, you should not use this facility. Affiliations will be numbered
% in order of appearance and this is the preferred way.
\icmlsetsymbol{equal}{*}

\begin{icmlauthorlist}
\icmlauthor{Nutan Chen}{vw}
\icmlauthor{Alexej Klushyn}{vw}
\icmlauthor{Francesco Ferroni}{aid}
\icmlauthor{Justin Bayer}{vw}
\icmlauthor{Patrick van der Smagt}{vw}
\end{icmlauthorlist}

\icmlaffiliation{vw}{Machine Learning Research Lab, Volkswagen Group, Munich, Germany}
\icmlaffiliation{aid}{Autonomous Intelligent Driving GmbH, Munich, Germany}

\icmlcorrespondingauthor{Nutan Chen}{nutan.chen@gmail.com}
%\icmlcorrespondingauthor{Eee Pppp}{ep@eden.co.uk}

% You may provide any keywords that you
% find helpful for describing your paper; these are used to populate
% the "keywords" metadata in the PDF but will not be shown in the document
\icmlkeywords{deep generative model, similarity, distance function, Riemannian manifold}

\vskip 0.3in
]

% this must go after the closing bracket ] following \twocolumn[ ...

% This command actually creates the footnote in the first column
% listing the affiliations and the copyright notice.
% The command takes one argument, which is text to display at the start of the footnote.
% The \icmlEqualContribution command is standard text for equal contribution.
% Remove it (just {}) if you do not need this facility.

\printAffiliationsAndNotice{}  % leave blank if no need to mention equal contribution
%\printAffiliationsAndNotice{\icmlEqualContribution} % otherwise use the standard text.

\begin{abstract}
%!TEX root = ../FMVAEs.tex

Measuring the similarity between data points often requires domain knowledge, which can in parts be compensated by relying on unsupervised methods such as latent-variable models, where similarity/distance is estimated in a more compact latent space.
Prevalent is the use of the Euclidean metric, which has the drawback of ignoring information about similarity of data stored in the decoder, as captured by the framework of Riemannian geometry.
We propose an extension to the framework of variational auto-encoders allows learning \emph{flat latent manifolds}, where the Euclidean metric is a proxy for the similarity between data points.
This is achieved by defining the latent space as a Riemannian manifold and by regularising the metric tensor to be a scaled identity matrix.
Additionally, we replace the compact prior typically used in variational auto-encoders with a recently presented, more expressive hierarchical one---and formulate the learning problem as a constrained optimisation problem.
We evaluate our method on a range of data-sets, including a video-tracking benchmark, where the performance of our unsupervised approach nears that of state-of-the-art supervised approaches, while retaining the computational efficiency of straight-line-based approaches.

\end{abstract}

%!TEX root = ../FMVAEs.tex
\section{Introduction}

Measuring the distance between data points is a central ingredient of many data analysis and machine learning applications. Several kernel methods (KernelPCA \citep{scholkopf1997kernel}, KernelNMF \citep{li2006relationships}, etc.), and other non-parametric approaches such as k-nearest neighbours \citep{altman1992introduction} rely on the availability of a suitable distance function.
Computer vision pipelines, e.g. tracking over time, perform matching based on similarity scores.

But designing a distance function can be challenging:
it is not always obvious to write down mathematical formulae that accurately express a notion of similarity.
Learning such functions has hence been proven as a viable alternative to manual engineering in this respect (NCA \citep{goldberger2005neighbourhood}, metric learning \citep{xing2003distance}, etc.).
Often, these methods rely on the availability of pairs labelled as similar or dissimilar.
A different route is that of exploiting the structure that latent-variable models learn.
The assumption that a set of high-dimensional observations is explained by points in a much simpler latent space underpins these approaches.
In their respective probabilistic versions, a latent prior distribution is transformed non-linearly to give rise to a distribution of observations.
The hope is that simple distances, such as the Euclidean distance measured in latent space, implement a function of similarity.
Yet, these approaches do not incorporate the variation of the observations with respect to the latent points.
For example, the observations will vary much more when a path in latent space will cross a class boundary. 

In fact, recent approaches to non-linear latent variable models, such as the generative adversarial network \citep{goodfellow2014generative} or the variational auto-encoder (VAE) \citep{KingmaW14, rezende14}, regularise the latent space to be compact, i.e. to remove low-density regions.
This is in contrast to the aforementioned hope that Euclidean distances appropriately reflect similarity.

The above insight leads us to the development of \emph{flat manifold} variational auto-encoders.
This class of VAEs defines the latent space as Riemannian manifold and regularises the Riemannian metric tensor to be a scaled identity matrix.
In this context, a \emph{flat manifold} is a Riemannian manifold, which is isometric to the Euclidean space.
To not compromise the expressiveness, we relax the compactness assumption and make use of a recently introduced hierarchical prior \citep{vahiprior2019}.
As a consequence, the model is capable of learning a latent representation, where the Euclidean metric is a proxy for the similarity between data points.
This results in a computational efficient distance metric which is practical for applications in real-time scenarios.

%!TEX root = ../FMVAEs.tex
\section{Variational Auto-Encoders with Flat Latent Manifolds}

\subsection{Background on Learning Hierarchical Priors in~VAEs}
\label{sec:methods-vae}
Latent-variable models are defined as
\begin{align}
\label{eq:nlvm}
p(\x) = \int p(\x|\z)\, p(\z)\, \mathrm{d}\z,
\end{align}
where $\z \in \mathbb{R}^{N_{z}}$ represents latent variables and $\x \in \mathbb{R}^{N_{x}}$ the observable data.  
The integral in Eq.~(\ref{eq:nlvm}) is usually intractable but it can be approximated by maximising the evidence lower bound (ELBO) \citep{KingmaW14, rezende14}:
\begin{align}
\label{eq:ELBO}
	\mathbb{E}_{p_\mathcal{D}(\mathbf{x})}\big[\log p_{\theta}(\mathbf{x})\big] \geq
	&\mathop{\mathbb{E}_{p_\mathcal{D}(\mathbf{x})}}\Big[\mathbb{E}_{q_\phi(\mathbf{z}\vert\mathbf{x})}\big[\log 
	p_\theta(\mathbf{x}\vert\mathbf{z})\big] \nonumber \\
	&-\mathop{\mathbb{KL}}\big(q_\phi(\mathbf{z}\vert\mathbf{x})\|\,p(\mathbf{z})\big)\Big],
\end{align}
where ${p_\mathcal{D}(\mathbf{x})=\frac{1}{N}\sum_{i=1}^{N}\delta(\mathbf{x}-\mathbf{x}_i)}$ is the empirical distribution of the data $\mathcal{D} = \{\mathbf{x_i}\}^N_{i=1}$.
The distribution parameters of the approximate posterior $q_{\phi}(\z|\x)$ and the likelihood $
p_\theta(\x|\z)$ are represented by neural networks.
The prior $p(\z)$ is usually defined as a standard normal distribution.
This model is commonly referred to as the variational auto-encoder (VAE).

However, a standard normal prior often leads to an over-regularisation of the approximate posterior, which results in a less informative learned latent representation of the data \citep{TomczakW18, vahiprior2019}.
To enable the model to learn an informative latent representation, \citet{vahiprior2019} propose to use a flexible hierarchical prior $p_\Theta(\mathbf{z}) = \int\! p_\Theta(\mathbf{z}\vert \mathbf{\zeta})\, p(\mathbf{\zeta})\, \mathrm{d}\mathbf{\zeta}$, where $p(\mathbf{\zeta})$ is the standard normal distribution.
Since the optimal prior is the aggregated posterior \citep{TomczakW18}, the above integral is approximated by an importance-weighted (IW) bound \citep{BurdaGS15} based on samples from $q_{\phi}(\z|\x)$.
This leads to a model with two stochastic layers and the following upper bound on the KL term: 
\begin{align}
\label{eq:vhp_kl_bound}
	\mathbb{E}&_{p_\mathcal{D}(\mathbf{x})}\mathop{\mathbb{KL}}\big(q_\phi(\mathbf{z}\vert\mathbf{x})\|\,p(\mathbf{z})\big)
	\leq 
	\mathcal{F}(\phi, \Theta, \Phi)
	\nonumber
	\\
	&\equiv
	\mathbb{E}_{p_\mathcal{D}(\mathbf{x})} \mathop{\mathbb{E}_{q_\phi(\mathbf{z}|\mathbf{x})}}\bigg[
	\log q_\phi(\mathbf{z}\vert\mathbf{x}) \nonumber \\
	& \quad -\mathop{\mathbb{E}_{\mathbf{\zeta}_{1:K}\sim q_\Phi(\mathbf{\zeta}|\mathbf{z})}}\Big[\log\frac{1}{K}\sum_{i=1}^{K}\frac{p_\Theta(\mathbf{z}\vert\mathbf{\zeta}_i)\, p(\mathbf{\zeta}_i)}{q_\Phi(\mathbf{\zeta}_i\vert\mathbf{z})}\Big]\bigg],
\end{align}
where $K$ is the number of importance samples.
Since it has been shown that high ELBO values do not necessarily correlate with informative latent representations \citep{2017arXiv171100464A, higgins2017beta}---which is also the case for hierarchical models \citep{sonderby2016}---different optimisation approaches have been introduced \citep{K16-1002, sonderby2016}.
\citet{vahiprior2019} follow the line of argument in \citep{rezende2018taming} and reformulate the resulting ELBO as the Lagrangian of a constrained optimisation problem:
\begin{align}
\label{eq:vhp_loss}
	\mathcal{L}&_\text{VHP}(\theta, \phi, \Theta, \Phi; \lambda) \equiv \nonumber\\
	&\mathcal{F}(\phi, \Theta, \Phi) + \lambda \big( \mathop{\mathbb{E}_{p_\mathcal{D}(\mathbf{x})}} \mathbb{E}_{q_\phi(\mathbf{z}\vert\mathbf{x})}\big[\text{C}_\theta(\mathbf{x}, \mathbf{z})\big]  
	- \kappa^2 \big), 
\end{align}
with the optimisation objective $\mathcal{F}(\phi, \Theta, \Phi)$, the inequality constraint {$\mathop{\mathbb{E}_{p_\mathcal{D}(\mathbf{x})}} \mathbb{E}_{q_\phi(\mathbf{z}\vert\mathbf{x})}\big[\text{C}_\theta(\mathbf{x}, \mathbf{z})\big]\leq \kappa^2$}, and the Lagrange multiplier $\lambda$.
$\text{C}_\theta(\mathbf{x}, \mathbf{z})$ is defined as the reconstruction-error-related term in $-\log p_\theta(\mathbf{x}\vert\mathbf{z})$.
Thus, we obtain the following optimisation problem:
\begin{align}
	\min_{\Theta, \Phi} \min_{\theta} \max_{\lambda} \min_{\phi} & \: \mathcal{L}_\text{VHP}(\theta, \phi, \Theta, \Phi; \lambda)  \quad \text{s.t.} \quad \lambda \geq 0.
\end{align}
Building on that, the authors propose an optimisation algorithm---including a $\lambda$-update scheme---to achieve a tight lower bound on the log-likelihood.
This approach is referred to as variational hierarchical prior (VHP) VAE.

\subsection{Learning Flat Latent Manifolds with VAEs}
\label{sec:rgvaes}
The VHP-VAE is able to learn a latent representation that corresponds to the topology of the data manifold \cite{vahiprior2019}.
However, it is not guaranteed that the (Euclidean) distance between encoded data in the latent space is a sufficient distance metric in relation to the observation space.
In this work, we aim to measure the distance/difference of observed data directly in the latent space by means of the Euclidean distance of the encodings.

\citet{ChenKK2018metrics, arvanitidis2017latentICLR} define the latent space of a VAE as a Riemannian manifold.
This approach allows for computing the observation-space length of a trajectory ${\gamma:[0, 1]\rightarrow\mathbb{R}^{N_{\z}}}$ in the latent space:
\begin{align} 
\label{eq:riemannian_distance}
L(\gamma) &= \int_0^1 \sqrt{\dot{\gamma}(t)^T\, \G\big(\gamma(t)\big)\: \dot{\gamma}(t) } \,\mathrm{d}t,
\end{align}
where $\G\in \mathbb{R}^{N_\z  \times N_\z}$ is the Riemannian metric tensor, and $\dot{\gamma}(t)$ the time derivative.
We define the \emph{observation-space distance} as the shortest possible path
\begin{align}
D=\min_\gamma L(\gamma)
\end{align}
between two data points.
The trajectory ${\gamma=\argmin_\gamma L(\gamma)}$ that minimises $L(\gamma)$ is referred to as the (minimising) geodesic.
In the context of VAEs, $\gamma$ is transformed by a continuous function $f(\gamma(t))$---the decoder---to the observation space. 
The metric tensor is defined as $\G(\z)=\J(\z)^{T}\J(\z)$, where $\J$ is the Jacobian of the decoder.

To measure the \emph{observation-space distance} directly in the latent space, distances in the observation space should be proportional to distances in the latent space:
\begin{align}
D \propto \, \| \z(1)-\z(0) \|_2,
\end{align}
where we define the Euclidean distance as the distance metric.
This requires that the Riemannian metric tensor is $\G\propto\mathds{1}$.
As a consequence, the Euclidean distance in the latent space corresponds to the \emph{observation-space distance}.
We refer to a manifold with this property as \emph{flat manifold} \citep{lee2006riemannian}.
To obtain a \emph{flat latent manifold}, the model typically needs to learn complex latent representations of the data (see experiments in Sec.~\ref{sec:results}).
Therefore, we propose the following approach:
(i) to enable our model to learn complex latent representations, we apply a flexible prior (VHP), which is learned by the model (empirical Bayes);
and (ii)~we regularise the curvature of the decoder such that~$\G\propto\mathds{1}$.

For this purpose, the VHP-VAE, introduced in Sec.~\ref{sec:methods-vae}, is extended by a Jacobian-regularisation term.
We define the regularisation term as part of the optimisation objective, which is in line with the constrained optimisation setting.
The resulting objective function is
\begin{align}
\label{eq:loss_no_aug}
	\mathcal{L}(\theta, \phi, \Theta, &\Phi; \lambda, \eta, c^2) = \mathcal{L}_\text{VHP}(\theta, \phi, \Theta, \Phi; \lambda)~+ 
	\nonumber	
	\\
	&~\eta \big( \mathop{\mathbb{E}_{p_\mathcal{D}(\mathbf{x})}} \mathbb{E}_{q_\phi(\mathbf{z}\vert\mathbf{x})}\big[\|\G(\z) - c^2\mathds{1}\|_2^2 \big]\big),	
\end{align}
where $\eta$ is a hyper-parameter determining the influence of the regularisation and $c$ the scaling factor.
Additionally, we use a stochastic approximation (first order Taylor expansion) of the Jacobian to improve the computational efficiency \citep{rifai2011higher}:
\begin{align}
\J_t(\z) = \lim_{\sigma \rightarrow 0 }\frac{1}{\sigma} \mathop{\mathbb{E}_{\epsilon \sim \mathcal{N} (0, \sigma^2)}} \big[f(\z +
\epsilon\,e_t) - f(\z)\big],
\end{align}
where $\J_t\in \mathbb{R}^{N_\x}$ is the Jacobian of the $t$-th latent dimension and $e_t$ a standard basis vector.
This approximation method allows for a faster computation of the gradient and avoids the second-derivative problem of piece-wise linear layers \citep{ChenKK2018metrics}.

However, the influence of the regularisation term in Eq.~(\ref{eq:loss_no_aug}) on the decoder function is limited to regions where data is available.
To overcome this issue, we propose to use \textit{mixup}, a data-augmentation method \citep{zhang2018mixup}, which was introduced in the context of supervised learning.
We extend this method to the VAE framework (unsupervised learning) by applying it to encoded data in the latent space.
This approach allows augmenting data by interpolating between two encoded data points~$\z_i$~and~$\z_j$: 
\begin{align}
\label{eq:mixup}
g(\z_i, \z_j) = (1-\alpha)\,\z_i + \alpha \,\z_j,
\end{align} 
with $\mathbf{x}_i, \mathbf{x}_j \sim p_\mathcal{D}(\mathbf{x}), \, \z_i \sim q_\phi(\mathbf{z}\vert\mathbf{x}_i), \, \z_j \sim q_\phi(\mathbf{z}\vert\mathbf{x}_j)$, and $\alpha \sim U(-\alpha_0, 1+\alpha_0)$. 
In contrast to \citep{zhang2018mixup}, where $\alpha \in [0, 1]$ limits the data augmentation to only convex combinations, we define $\alpha_0 > 0$ to take into account the outer edge of the data manifold.
By combining \emph{mixup}~in~Eq.~(\ref{eq:mixup}) with Eq.~(\ref{eq:loss_no_aug}), we obtain the objective function of our \emph{flat manifold} VAE (FMVAE):
\begin{align}
\label{eq:loss}
	&\mathcal{L}_\text{\methodreg}(\theta, \phi, \Theta, \Phi; \lambda, \eta, c^2) =\mathcal{L}_\text{VHP}(\theta, \phi, \Theta, \Phi; \lambda)~ +
	\nonumber
	\\
	&\quad\eta \,\mathop{\mathbb{E}_{\mathbf{x}_{i,j} \sim p_\mathcal{D}(\mathbf{x})}} \mathbb{E}_{\z_{i,j} \sim q_\phi(\mathbf{z}\vert\mathbf{x}_{i,j})}  \big[ \|  \G(g(\z_i, \z_j))  - c^2\mathds{1}\|_2^2 \big].
\end{align}
Inspired by batch normalisation, we define the squared scaling factor to be the mean over the batch samples and diagonal elements of $\G$ (see App.~\ref{app:compc} for empirical support):
\begin{align}
	c^2=\frac{1}{N_\z}\,\mathop{\mathbb{E}_{\mathbf{x}_{i,j} \sim p_\mathcal{D}(\mathbf{x})}} \mathbb{E}_{\z_{i,j} \sim q_\phi(\mathbf{z}\vert\mathbf{x}_{i,j})}\big[\mathrm{tr}(\G(g(\z_i, \z_j)))\big].
\end{align}
The optimisation algorithm Alg.~\ref{alg:flm_rewo}, and further details about the optimisation process can be found in App.~\ref{app:alg}.

By using augmented data, we regularise $\G$ to be a scaled identity matrix for the \emph{entire} latent space enclosed by the data manifold.
As a consequence, the function $f(\z)$ (decoder) is---up to the scaling factor $c$---distance-preserving since $D_\x(f(\z_i), f(\z_j))\approx c \,D_\z(\z_i,\z_j)$, where $D_\x$ and $D_\z$ refer to the distance in the observation and latent space, respectively.

%\TODO{remove:}The decoder of the proposed approach satisfies the Lipschitz continuity condition $D_\x(f(\z_i),f(\z_j)) \le  a \, D_\z(\z_i,\z_j)$. We consider the decoder function, and hence the latent space as \emph{smooth} if $\exists \, c \leq a$, where $a$ is the Lipschitz constant.

%!TEX root = ../FMVAEs.tex
\section{Related Work}

\noindent{\textbf{Interpretation of the VAE's latent space}}.
In general, the latent space of VAEs is considered to be Euclidean \citep[e.g.][]{Kingma2016,higgins2017beta}, but it is not constrained to be Euclidean.
This can be problematic if we require a precise metric that is based on the latent space.
Some recent works \citep{mathieu2019hierarchical, grattarola2018learning} adapted the latent space to be non-Euclidean to match the data structure.
We solve the problem from another perspective: we enforce the latent space to be Euclidean.

\noindent{\textbf{Jacobian and Hessian regularisation}}.
In \citep{rifai2011manifold}, the authors proposed to regularise the Jacobian and Hessian of the encoder.
However, it is more difficult to augment data in the observation space than in the latent space.
In \citep{hadjeres2017glsr}, the Jacobian of the decoder was regularised to be as small as possible/zero.
On the contrary, we regularise the the Riemannian metric tensor to be a scaled identity matrix, and hence the Jacobian to be constant, and hence the Hessian to be zero.
\cite{Nie2019TowardsAB} regularised the Jacobian with respect to the weights for GANs.
In terms of supervised learning, \cite{jakubovitz2018} used Jacobian regularisation to improve the robustness for classification.

\noindent{\textbf{Metric learning}}.
Various metric learning approaches for both deep supervised and unsupervised models were proposed.
For instance, deep metric learning \citep{hoffer2015deep} used a triplet network for supervised learning.
\cite{karaletsos2016bayesian} introduced an unsupervised metric learning method, where a VAE is combined with triplets.
However, a human oracle is still required. 
By contrast, our approach is completely based on unsupervised learning, using the Euclidean distance in the latent space as a distance metric.
Our proposed method is similar to the metric learning methods such as Large Margin Nearest Neighbour \citep{weinberger2009distance}, which pulls target neighbours together and pushes impostors away. The difference is that our approach is an unsupervised method. 

\noindent{\textbf{Constraints in latent space}}.
Constraints on time \citep[e.g.][]{wang2007gaussian,chen2016dynamic, chen2015efficient} allow obtaining similar distance metrics in the latent space.
Additionally, due to the missing data between different sequence steps, constraints on time cannot guarantee that the metric is correct between different sequences. 
However, our method can be used for general data-sets.

\noindent{\textbf{Data augmentation}}.
The latent space is formed arbitrarily in regions where data is missing.
\citet{zhang2018mixup} proposed \textit{mixup}, an approach for augmenting data and labels for supervised learning.
Various follow-up studies of \textit{mixup} were developed, such as \citep{verma2018manifold, beckham2019adversarial}. \citep{verma2018manifold} considered \textit{mixup} of hidden representations of training data to flatten the class-specific state distribution.
We extend \textit{mixup} to the VAE framework (unsupervised learning) by applying it to encoded data in the latent space of generative models.
This facilitates the regularisation of regions where no data is available.
As a consequence, similarity of data points can be measured in the latent space by applying the Euclidean metric.

\noindent{\textbf{Geodesic}}.
Recent studies on geodesics for generative models \citep[e.g.][]{TosiHVL14, ChenKK2018metrics, arvanitidis2017latentICLR} are focusing on methods for computing/finding the geodesic in the latent space. 
By contrast, we use the geodesic/Riemannian distance for influencing the learned latent manifold.
\cite{frenzel2019latent} projected the latent space to a new latent space, where the geodesic is equivalent to the Euclidean interpolation.
However, these two separate processes---VAEs and projection---probably hinder the model to find the latent features autonomously. 
Another difference is the assumption of previous work is that distances, defined by geodesics, can only be measured by following the data manifold.
This is useful in scenarios such as avoiding unseen barriers between two data points, e.g., \cite{chen2018active}, but it does not allow measuring distances between different categories.
In this work, we focus on learning a general distance metric.

%!TEX root = ../FMVAEs.tex
\section{Experiments}
\label{sec:results}

We test our method on artificial pendulum images, human motion data, MNIST, and MOT16. 
We measure the performance in terms of equidistances, interpolation smoothness, and distance computation. Additionally, our method is applied to a real-world environment---a video-tracking benchmark.
Here, the tracking and re-identification capabilities are evaluated.

The Riemannian metric tensor has many intrinsic properties of a manifold and measures local angles, length, surface area, and volumes \citep{bronstein2017geometric}.
Therefore, the models are quantified based on the Riemannian metric tensor by computing condition numbers and magnification factors.
The condition number, which shows the ratio of the most elongated to the least elongated direction, is defined as $k(\G) = \frac{S_\text{max}(\G)}{S_\text{min}(\G)}$, where $S_\text{max}$ is the largest eigenvalue of $\G$.
The magnification factor $\MF(\z)\equiv \sqrt{\det\G(\z)}$ \citep{bishop1997magnification} depicts the sensitivity of the likelihood functions.
When projecting from the Riemannian (latent) to the Euclidean (observation) space, the $\MF$ can be considered a scaling coefficient.
Since we cannot directly compare the $\MF$s of different models, the $\MF$s are normalised/divided by their means.
The closer the conditional number and the normalised MF are to one, the more invariant is the model with respect to the Riemannian metric tensor.
In other words: the conditional number and the normalised MF are metrics to evaluate whether $\G(\z)$ is approximately constant and proportional to $\mathds{1}$.

In order to make the visualisations of the magnification factor in Sec.~\ref{sec:pendulum} (Fig.~\ref{fig:pendulum_equal}) and  Sec.~\ref{sec:human} (Fig.~\ref{fig:human_equal} \& Fig.~\ref{fig:human_comparison}) comparable, we define the respective upper range of the colour-bar as $\frac{\text{max}(\MF_\text{VAE-VHP} (\text{grid\_area})) \cdot \text{mean}(\MF_\text{\methodreg}(\text{data}))} { \text{mean}(\MF_\text{\methodreg}(\text{data}))}$. 
$\MF(\text{data})$ and $\MF(\text{grid\_area})$ are computed with training data and by using a grid area, respectively.

To be in line with previous literature \citep[e.g.][]{higgins2017beta, sonderby2016}, we use the $\beta$-parametrisation of the Lagrange multiplier $\beta=\frac{1}{\lambda}$ in our experiments.

\subsection{Artificial Pendulum Data-set}
\label{sec:pendulum}

\begin{figure}[!ht]
    \centering
    \begin{subfigure}[b]{0.67\columnwidth}
      \centering
        \includegraphics[width=\textwidth]{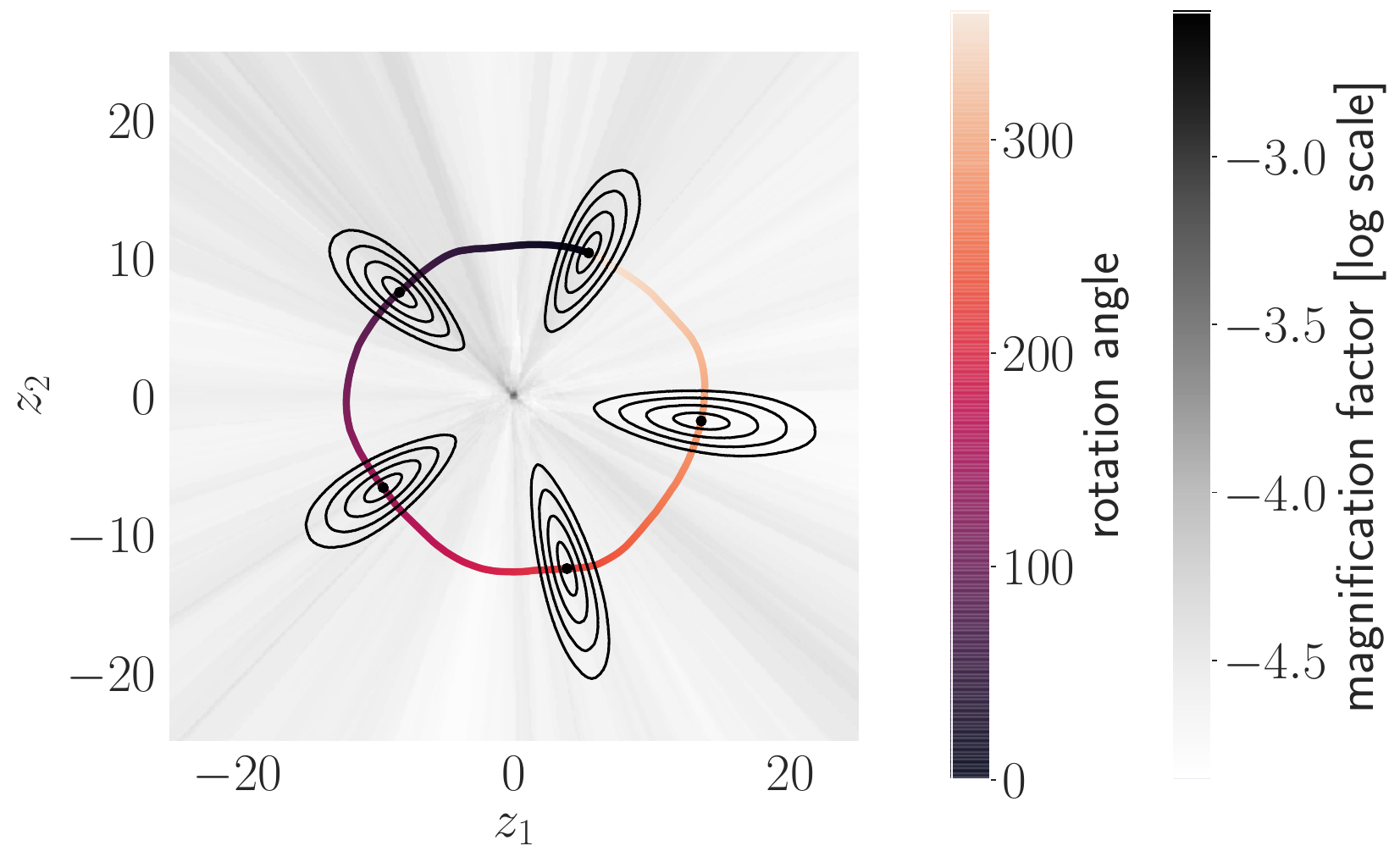}
        \caption{\methodreg}%
        \label{fig:pendulum_equal_reg}
    \end{subfigure}
    \\
    \begin{subfigure}[b]{0.67\columnwidth}
      \centering
        \includegraphics[width=\textwidth]{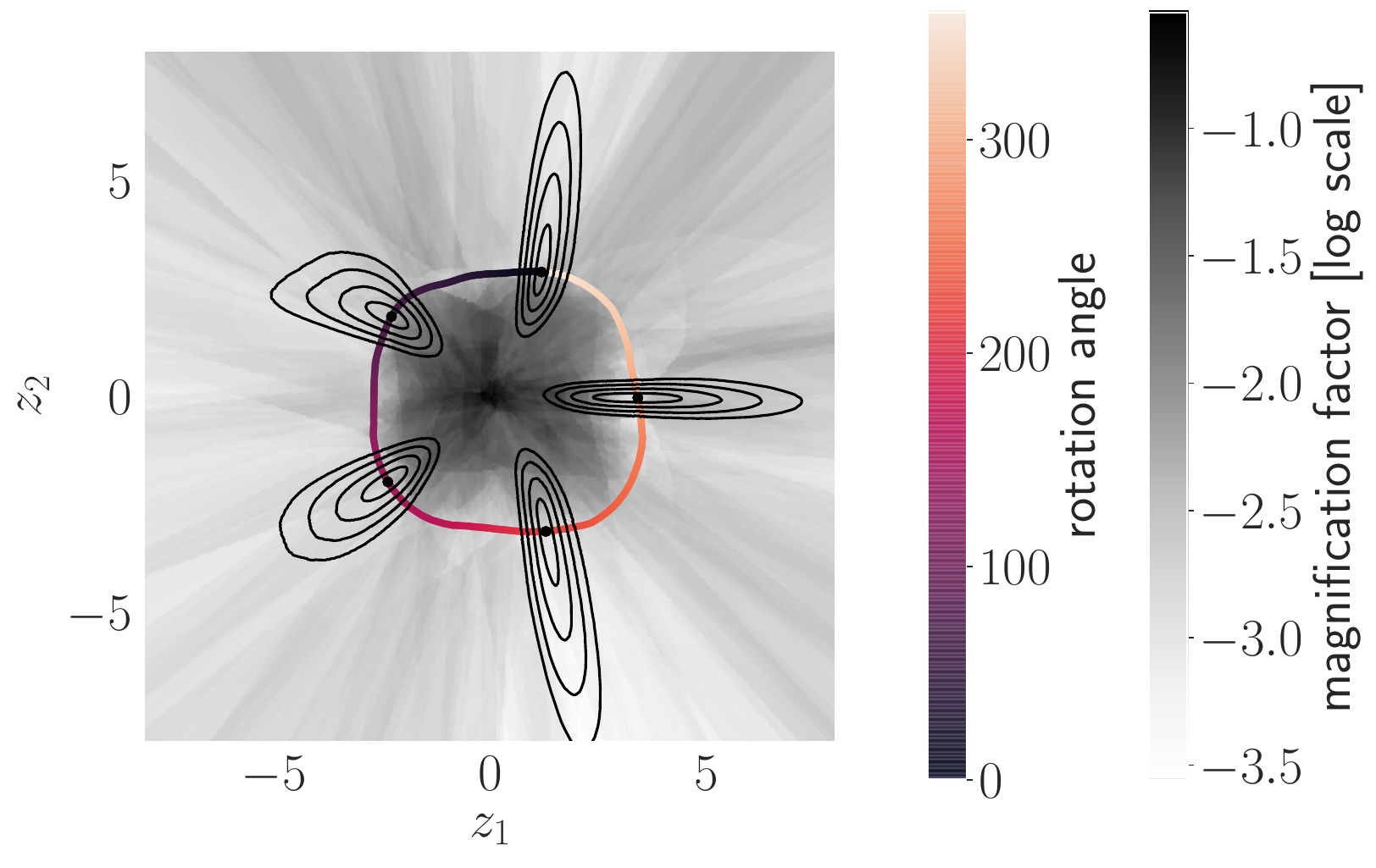}
        \caption{VHP-VAE}%
    \end{subfigure}
    \caption{Latent representation of pendulum data: the contour plots illustrate curves of equal \emph{observation-space distance} to the respective encoded data point. Distances are calculated using Eq.~(\ref{eq:riemannian_distance}). The grey-scale displays $\MF(\z)$. Note:~round, homogeneous contour plots indicate that $\G(\z)\propto\mathds{1}$.}
    \label{fig:pendulum_equal}
\end{figure}

\begin{figure}[!ht]
	\centering
	\includegraphics[width=0.14\textwidth]{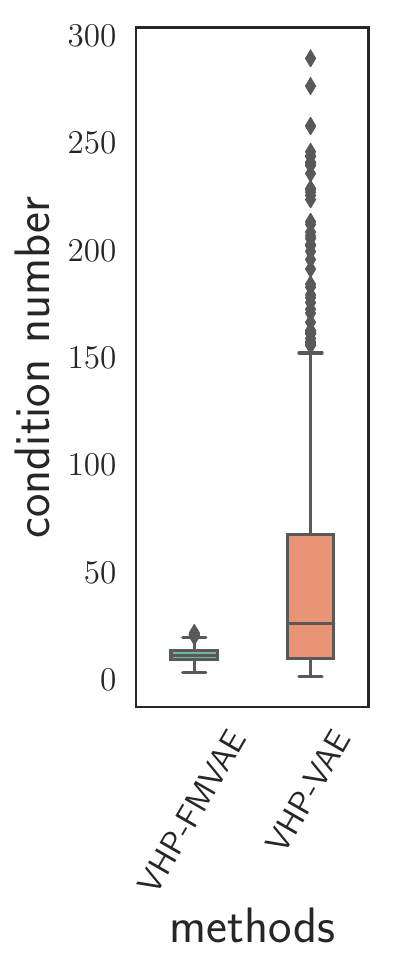}
	\includegraphics[width=0.14\textwidth]{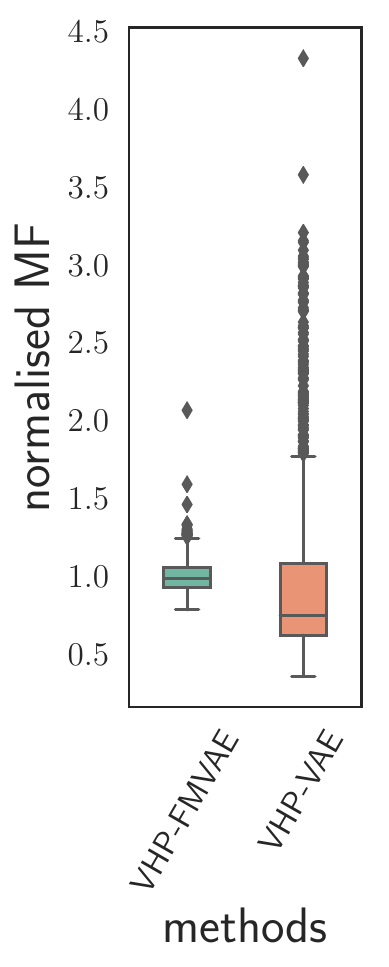}
	\caption{Pendulum data: if both the condition number and the normalised MF values are close to one, it indicates that $\G(\z)\propto\mathds{1}$. The box-plots are based on 1,000 generated samples.}%
	\label{fig:pendulum_cn_mf}
\end{figure}

The pendulum data-set \citep{vahiprior2019, ChenKK2018metrics} consists of $16{\times}16$-pixel images generated by a pendulum simulator. The pendulum has one degree of freedom, and the joint is located in the centres of the images.
We generated $15 \cdot10^3$ images with joint angles uniformly in the ranges of $[0, 360)$ degrees.
Additionally, we added $0.05$ Gaussian noise to each pixel.

As seen in Fig.~\ref{fig:pendulum_equal}, without regularisation, the contour lines are denser in the centre of the latent space. 
The reason is that, in contrast to the VHP-VAE, the regularisation term in the \methodreg \emph{smoothens} the latent space ($\G\approx c\,\mathds{1}$)---visualised by the $\MF$ and the equidistance plots.
In Fig.~\ref{fig:pendulum_cn_mf}, \methodreg and VAE-VHP are compared in terms of condition number and normalised $\MF$.
In both cases the \methodreg outperforms the VHP-VAE.

\subsection{Human Motion Capture Database}
\label{sec:human}

To evaluate our approach on the CMU human motion data-set (\url{http://mocap.cs.cmu.edu}), we select five different movements: 
walking (subject 35), jogging (subject 35), balancing (subject 49), punching (subject 143), and kicking (subject 74).
After data pre-processing, the input data is a 50-dimensional vector of the joint angles.
Note that the data-set is not balanced: walking, for example, has more data points than jogging.

\begin{figure}[ht!]
    \centering
    \begin{subfigure}[b]{\columnwidth}
        \includegraphics[width=\textwidth]{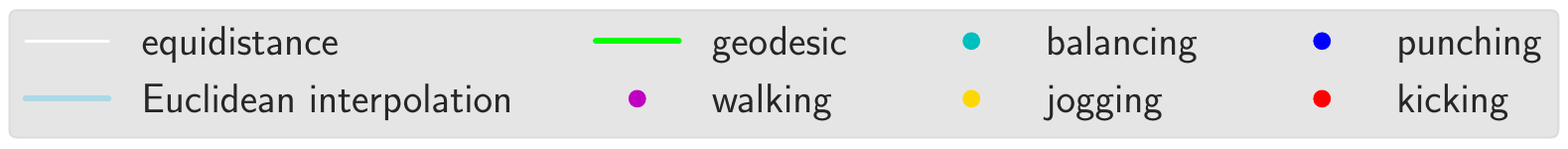}
        \label{fig:human_legend}
    \end{subfigure}\\
%    \vskip +0.1in
    \begin{subfigure}[b]{0.70\columnwidth}
        \includegraphics[width=\textwidth]{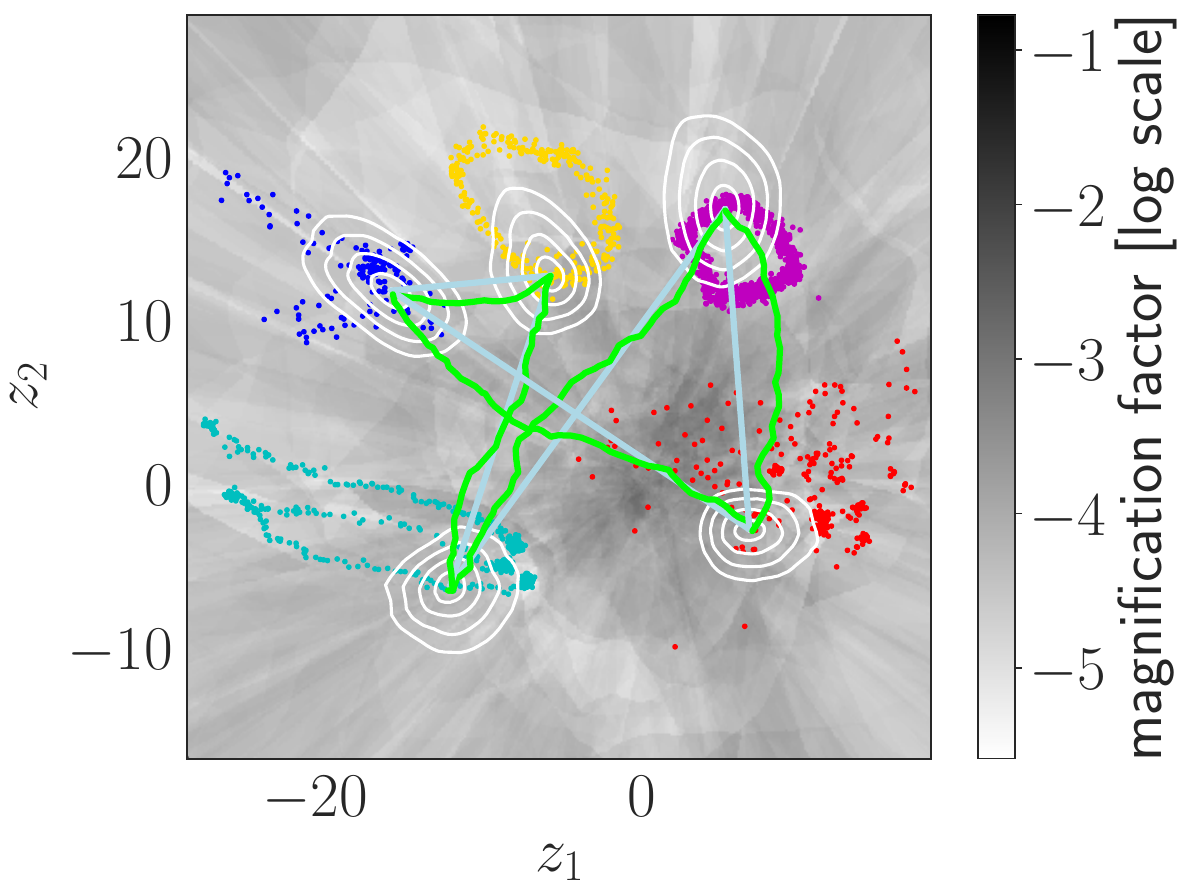}
        \caption{\methodreg}%
        \label{fig:human_equal_reg}
    \end{subfigure}
    \\
    \begin{subfigure}[b]{0.67\columnwidth}
        \includegraphics[width=\textwidth]{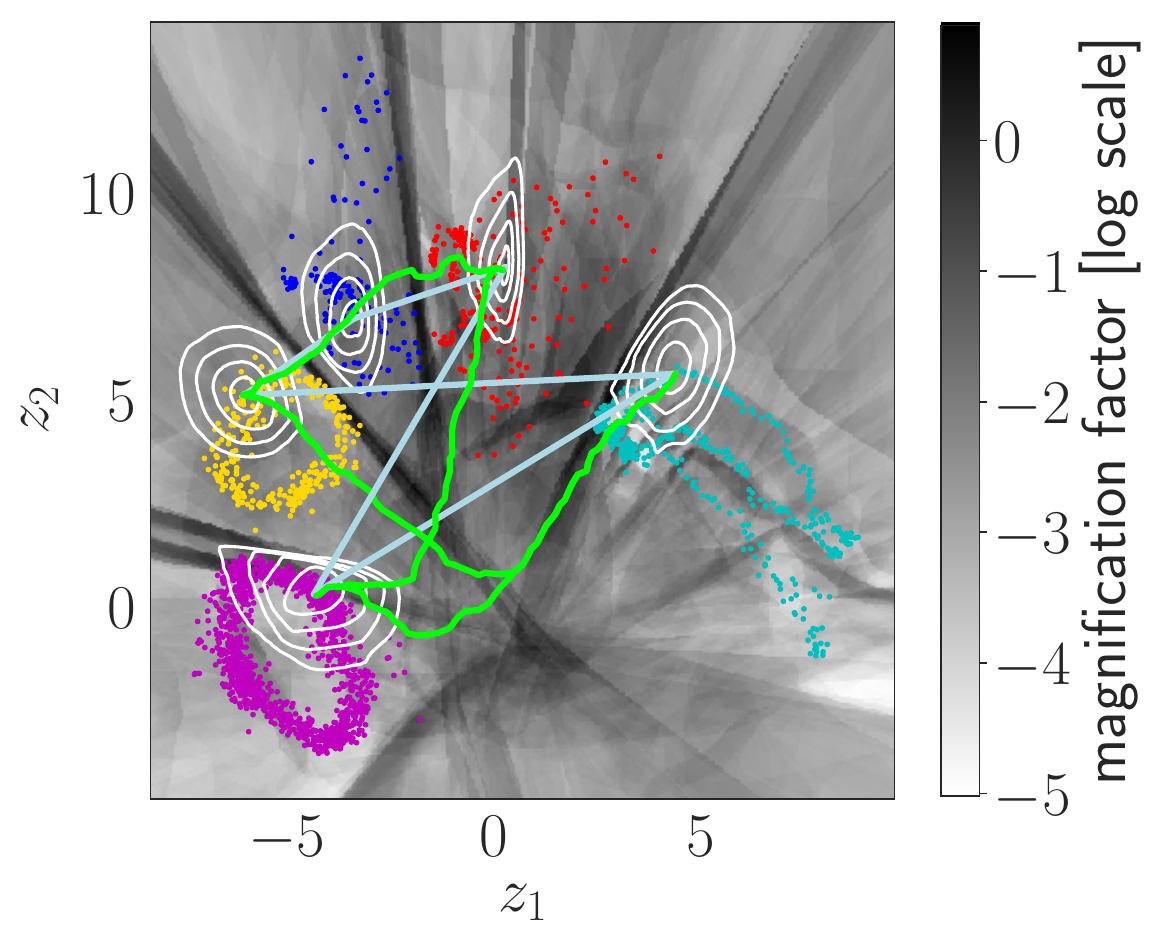}
        \caption{VAE-VHP}%
        \label{fig:human_equal_noreg}
    \end{subfigure}
    \caption{Latent representation of human motion data: the contour plots illustrate curves of equal \emph{observation-space distance} to the respective encoded data point. The grey-scale displays $\MF(\z)$. Note:~round, homogeneous contour plots indicate that ${\G(\z)\propto\mathds{1}}$. In case of the \methodreg (a), Jogging is a large-range movement compared with walking, so that jogging is reasonably distributed on a larger area in the latent space than walking. By contrast, in case of the VHP-VAE (b), the latent representation of walking is larger than the one of jogging. Additionally, geodesics are compared to the corresponding Euclidean interpolations. The Euclidean interpolations in (a) are much closer to the geodesics.}
    \label{fig:human_equal}
\end{figure}

\begin{figure}[ht!]
    \centering
    \begin{subfigure}[b]{0.40\columnwidth}
        \includegraphics[width=\textwidth]{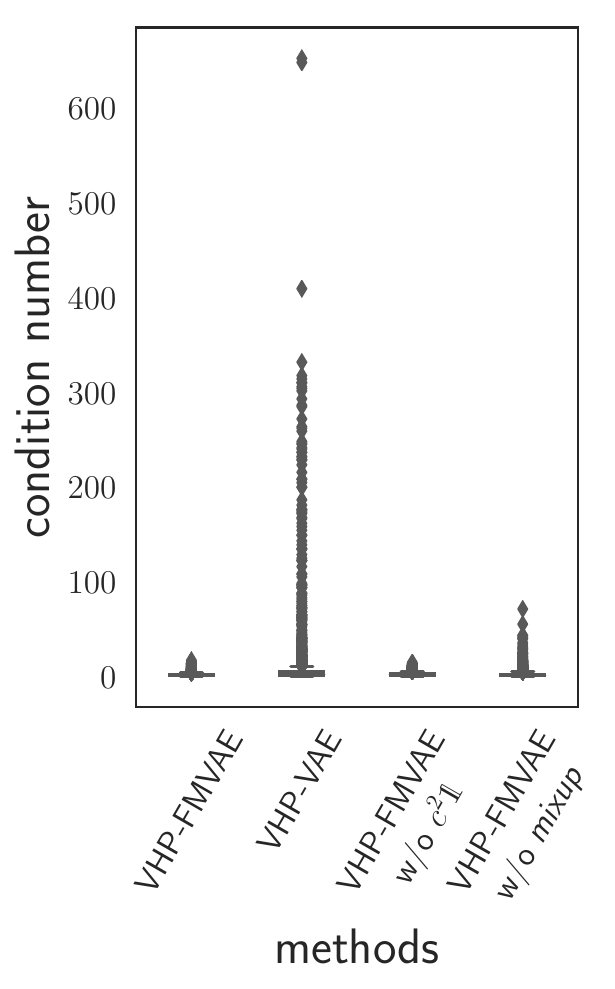}
        \label{fig:human_cn}
    \end{subfigure}
    %\hfill
    \begin{subfigure}[b]{0.40\columnwidth}
        \includegraphics[width=\textwidth]{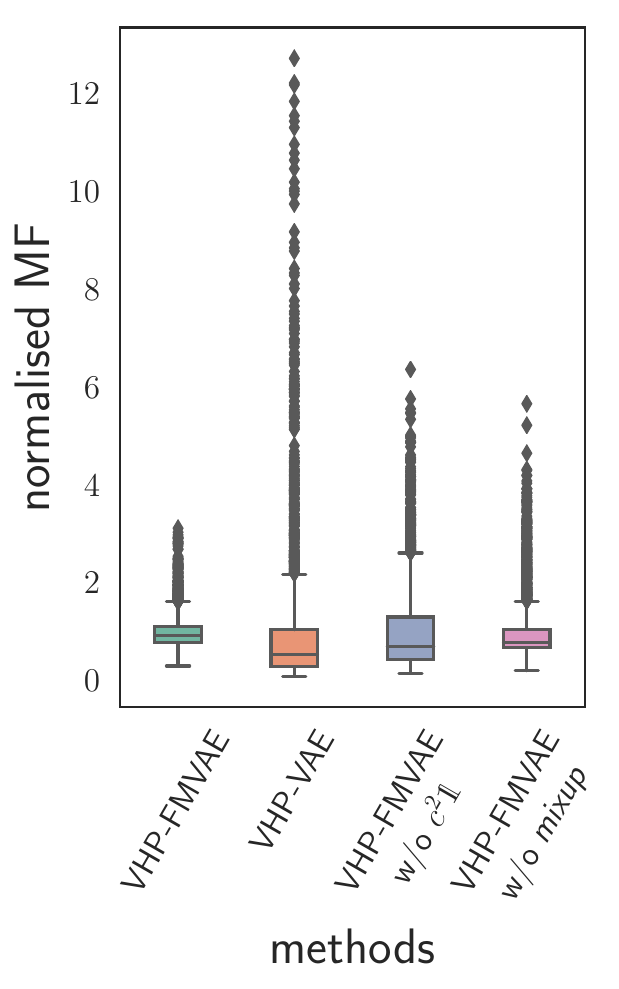}
        %\caption{Normalised MF.}%
        \label{fig:human_mf}
    \end{subfigure}
	\vskip -0.1in
    \caption{Human motion data: if both the condition number and the normalised MF values are close to one, it indicates that $\G(\z)\propto\mathds{1}$. The box-plots are based on 3,000 generated samples.}
    \label{fig:human_cn_mf}
\end{figure}

\begin{figure}[ht!]
    \centering
        \includegraphics[width=0.38\textwidth]{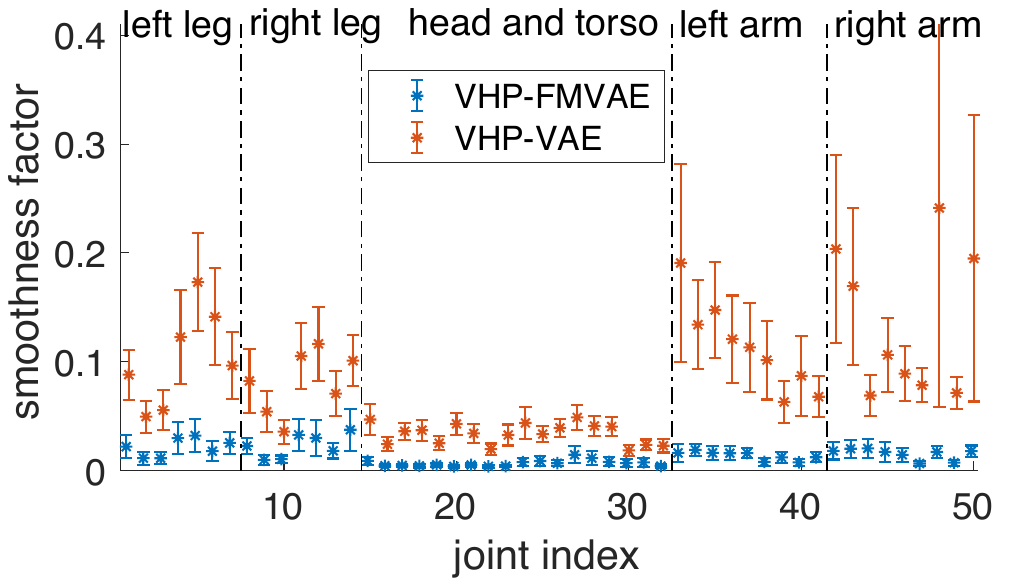}
    \caption{Smoothness measure of the human-movement interpolations. The mean and standard deviation are displayed for each joint: the smaller the value, the smoother the interpolation.
    }
    \label{fig:human_smoothness}
\end{figure}

\begin{figure}[ht!]
    \centering
    \begin{subfigure}[b]{0.495\columnwidth}
        \includegraphics[width=\textwidth]{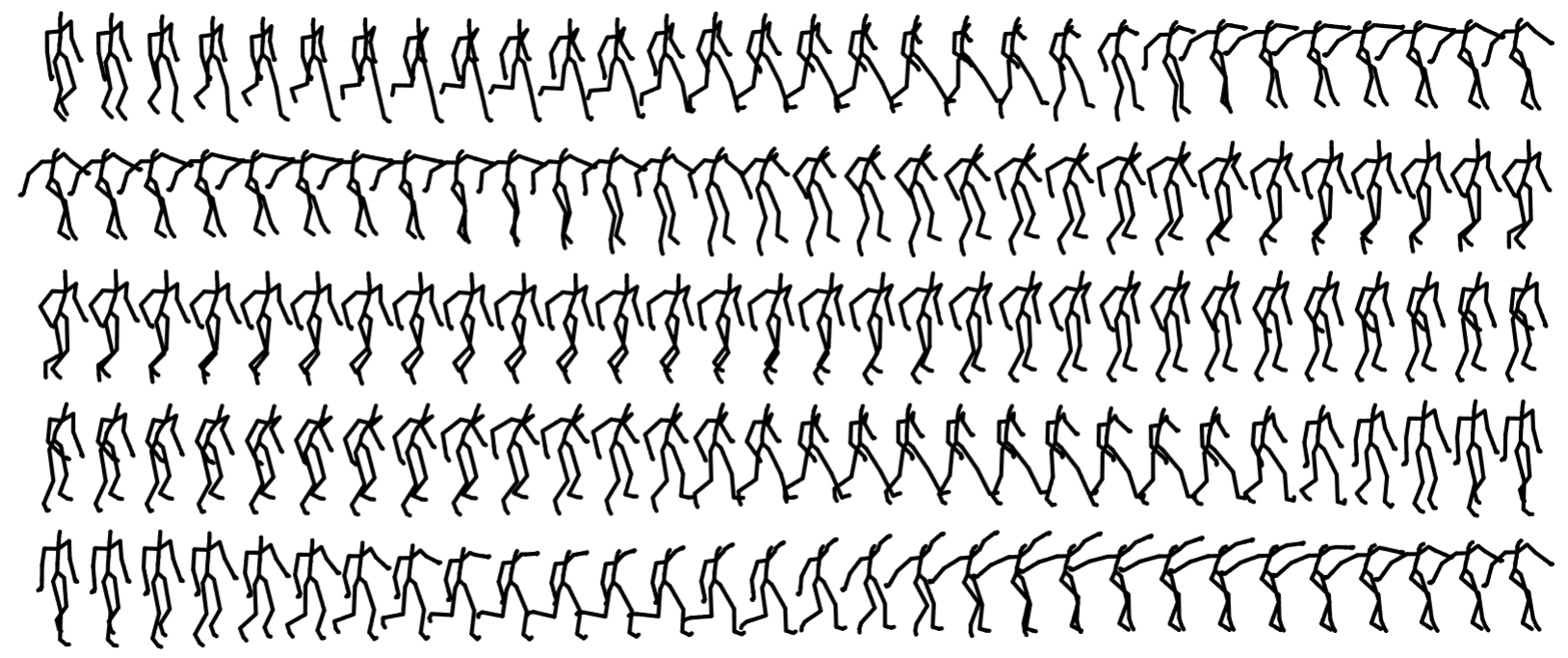}
        \caption{\methodreg}%
        \label{fig:human_inter_reg}
    \end{subfigure}
    \hfill
    \begin{subfigure}[b]{0.495\columnwidth}
        \includegraphics[width=\textwidth]{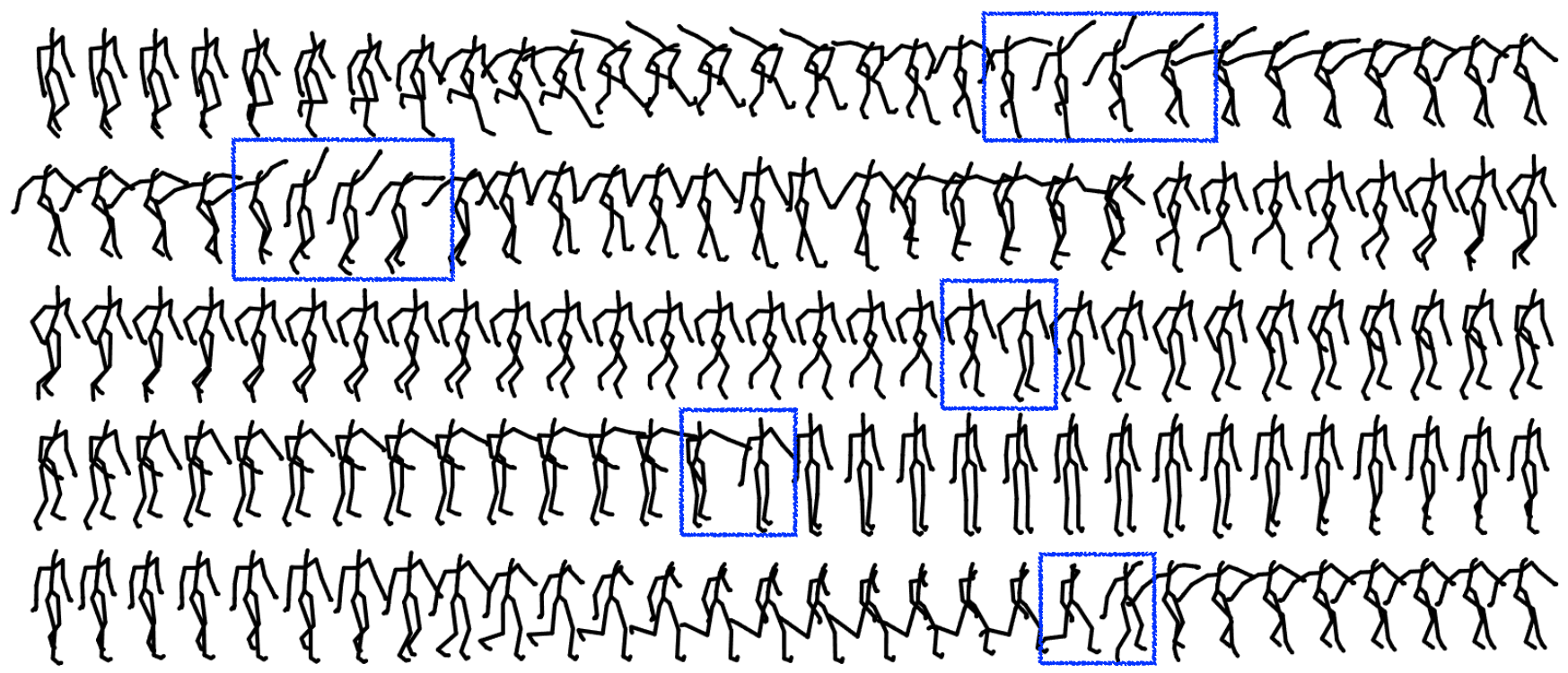}
        \caption{VHP-VAE}%
        \label{fig:human_interl_noreg}
    \end{subfigure}
    \caption{Human-movement reconstructions of Euclidean interpolations in the latent space. Discontinuities in the motions are marked by blue boxes.}
     \label{fig:human_interl}
\end{figure}

\begin{table}[ht!]
\vskip -0.1in
\caption{Verification of the distance metric. The table shows the length ratio of the Euclidean interpolation to the geodesic. Additionally, we list the ratio of the related distances in the observation space.}
\label{tab:geodesic}
\begin{center}
\begin{footnotesize}
\begin{sc}
%\resizebox{0.75\textwidth}{!}{%					
\begin{tabular}{llcc}
\toprule
data-set	& 	method	& observation & latent \\
\midrule
Human 	& \methodreg	& \textbf{1.02 $\pm$ 0.06} 	& \textbf{0.93 $\pm$ 0.03} \\ 
		& VHP-VAE	& 1.23 $\pm$ 0.20 			& 0.82 $\pm$ 0.10\\
\midrule
MNIST	& \methodreg	&  \textbf{1.01 $\pm$ 0.08} 	&  \textbf{0.92 $\pm$ 0.05}\\
		& VHP-VAE 	& 1.13 $\pm$0.22  			& 0.70 $\pm$ 0.31 \\
\bottomrule
\end{tabular}
%}
\end{sc}
\end{footnotesize}
\end{center}
\vskip -0.1in
\end{table}

\begin{figure}[ht!]
    \centering
    \begin{subfigure}[b]{\columnwidth}
        \centering
        \includegraphics[width=0.7\textwidth]{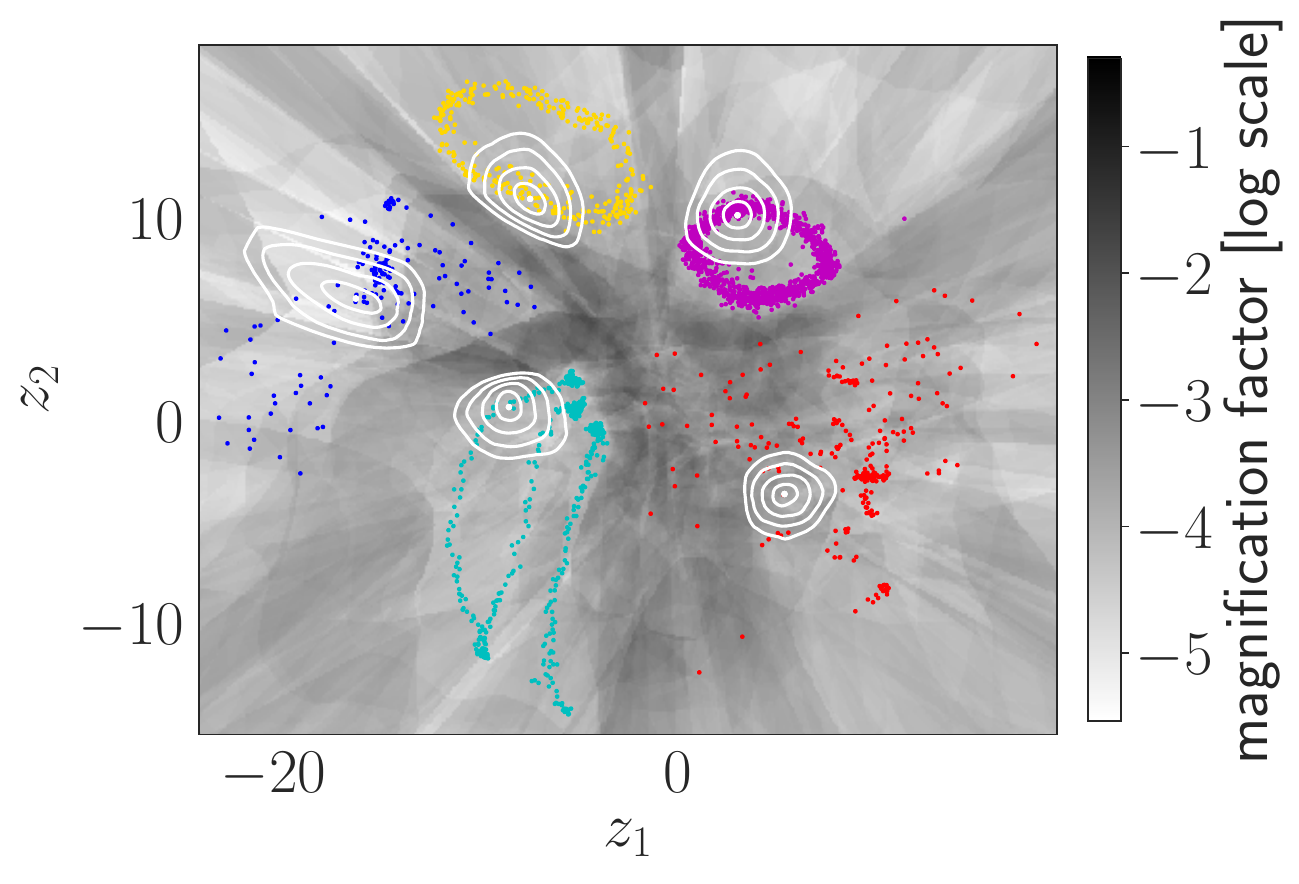}
        \caption{\methodreg without \textit{mixup}}
        \label{fig:human_no_augmentation}
    \end{subfigure}
    \\
    %\hfill
    \begin{subfigure}[b]{\columnwidth}
        \centering
        \includegraphics[width=0.68\textwidth]{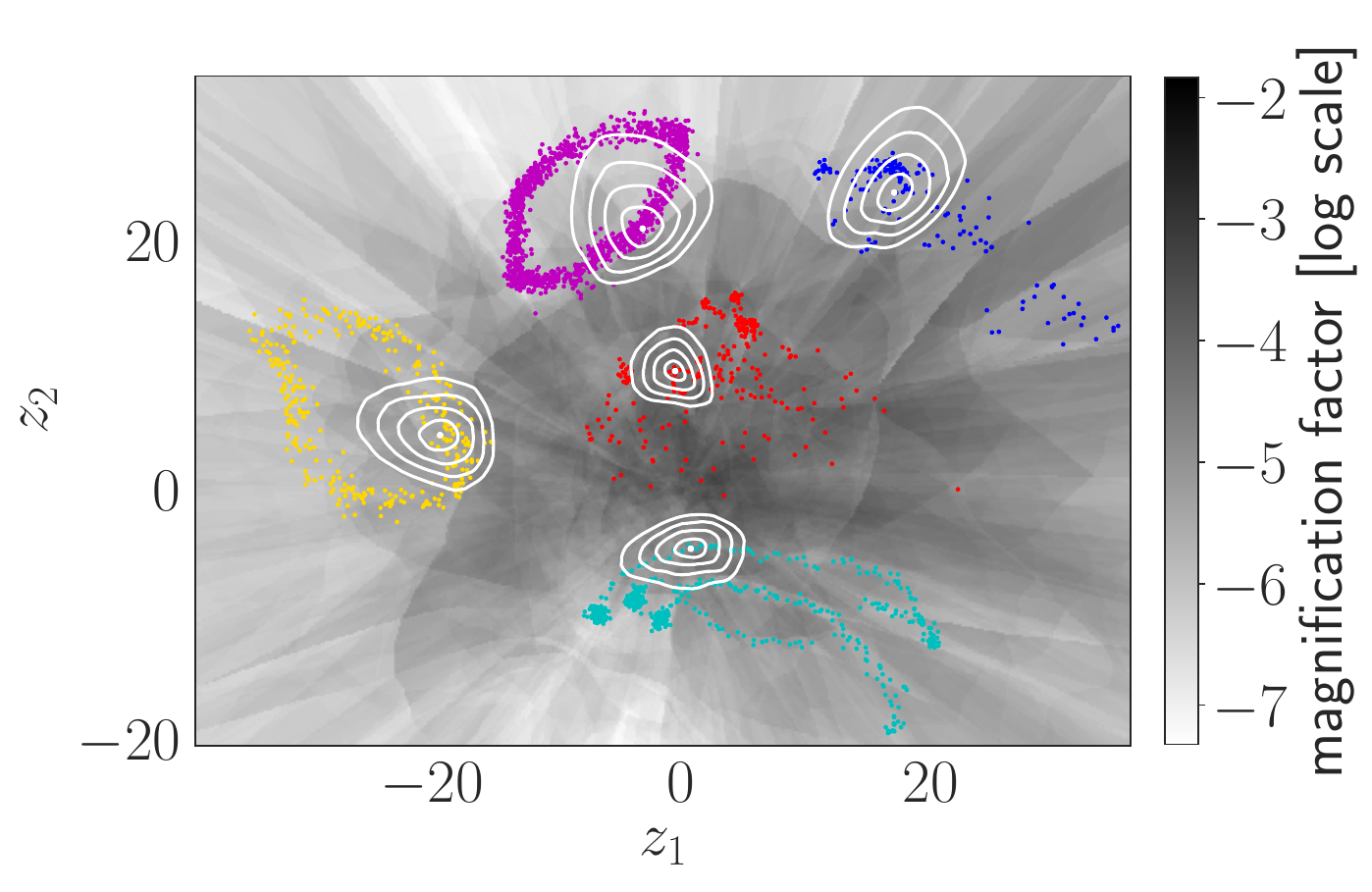}
        \caption{\methodreg without the identity term $c^2\mathds{1}$}
        \label{fig:human_no_normalisation}
    \end{subfigure}
    \caption{Influence of the data augmentation and the identity term $c^2\mathds{1}$ on the learned latent representation of human movement data. The movements are coloured as in Fig.~\ref{fig:human_equal}. (a) If not applying \emph{mixup}, regions, where data is missing (e.g., between two movements), have a high $\MF$ and distorted equidistance contours. (b) regularising the metric tensor, and hence the Jacobian to be zero, does not allow the model to learn a \emph{flat latent manifold}. The equidistance contours are scaled differently at various locations in the latent space. 
Without $c^2\mathds{1}$ term as in \citep{hadjeres2017glsr}, it cannot reduce the distance for points with high similarities. 
For instance, the walking is not squeezed as in Fig.~\ref{fig:human_equal_reg} in the latent space. Therefore, the walking is not distributed smaller than jogging.
}
    \label{fig:human_comparison}
\end{figure}

\noindent{\textbf{Equidistance plots}}.
In Fig.~\ref{fig:human_equal}, we randomly select a data point from each class as centres of the equidistance plots. 
In case of our proposed method, the equidistance plots are homogeneous, while in case of the VHP-VAE, the equidistance contour lines are distorted in regions of high $\MF$ values. 
Thus, the mapping from latent to observation space learned by the \methodreg is approximately distance preserving. 
Additionally, we use the condition number and the normalised $\MF$ to evaluate $\G$ based on 3,000 random samples.  
In contrast to the VHP-VAE, both the condition number and the normalised MF values of the \methodreg are close to one, which indicates that $\G(\z)\propto\mathds{1}$.

\noindent{\textbf{Smoothness}}.
We randomly sample 100 pair points and linearly interpolate between each pair.
The second derivative of each trajectory is defined as the smoothness factor. Fig.~\ref{fig:human_smoothness} illustrates that the \methodreg significantly outperforms the VAE-VHP in terms of smoothness. 
Fig.~\ref{fig:human_interl} shows five examples of the interpolated trajectories. 

\noindent{\textbf{Verification of the distance metric}}. 
To verify that the Euclidean distance in the latent space corresponds to the geodesic distance, we  approximates the geodesic by using a graph-based approach \citep{chen2019fast}. 
The graph of the baseline has 14,400 nodes, which are sampled in the latent space using a uniform distribution.
Each node has 12 neighbours. 
In Fig.~\ref{fig:human_equal}, five geodesics each are compared to the corresponding Euclidean interpolations.
Tab.~\ref{tab:geodesic} shows the ratios of Euclidean distances in latent space to geodesics distances, as well as the related ratios in the observation space.
To compute the ratios, we randomly sampled 100 pairs of points and interpolated between each pair.
If the ratio of the distances is close to one, the Euclidean interpolation approximates the geodesic. 
The \methodreg outperforms the VAE-VHP.
%As a side effect, the proposed method is 1,000 times faster than the graph-based method in terms of searching for the geodesics. 

\noindent{\textbf{Influence of the data augmentation and the identity term $c^2\mathds{1}$}}.
Fig.~\ref{fig:human_cn_mf} and Fig.~\ref{fig:human_no_augmentation} show the influence of the data augmentation (see Sec.~\ref{sec:rgvaes}).
Without data augmentation, the influence of the regularisation term is limited to regions where data is available, as verified by the high $\MF$ values between the different movements.
As an additional experiment, Fig.~\ref{fig:human_cn_mf} and Fig.~\ref{fig:human_no_normalisation} illustrates the influence of the identity term $c^2\mathds{1}$. 
If we remove it, the regularisation term becomes  $\|\G(g(\z_i, \z_j))\|_2^2$.
As a consequence, 
the model is not able to learn a \emph{flat latent manifold}.

\subsection{MNIST}
\label{sec:mnist}

The binarised MNIST data-set \citep{larochelle2011neural} consists of 50,000 training and 10,000 test images of handwritten digits (zero to nine) with $28{\times}28$ pixels in size.

\begin{figure}[ht!]
    \centering
    \begin{subfigure}[b]{0.49\columnwidth}
        \includegraphics[width=\textwidth]{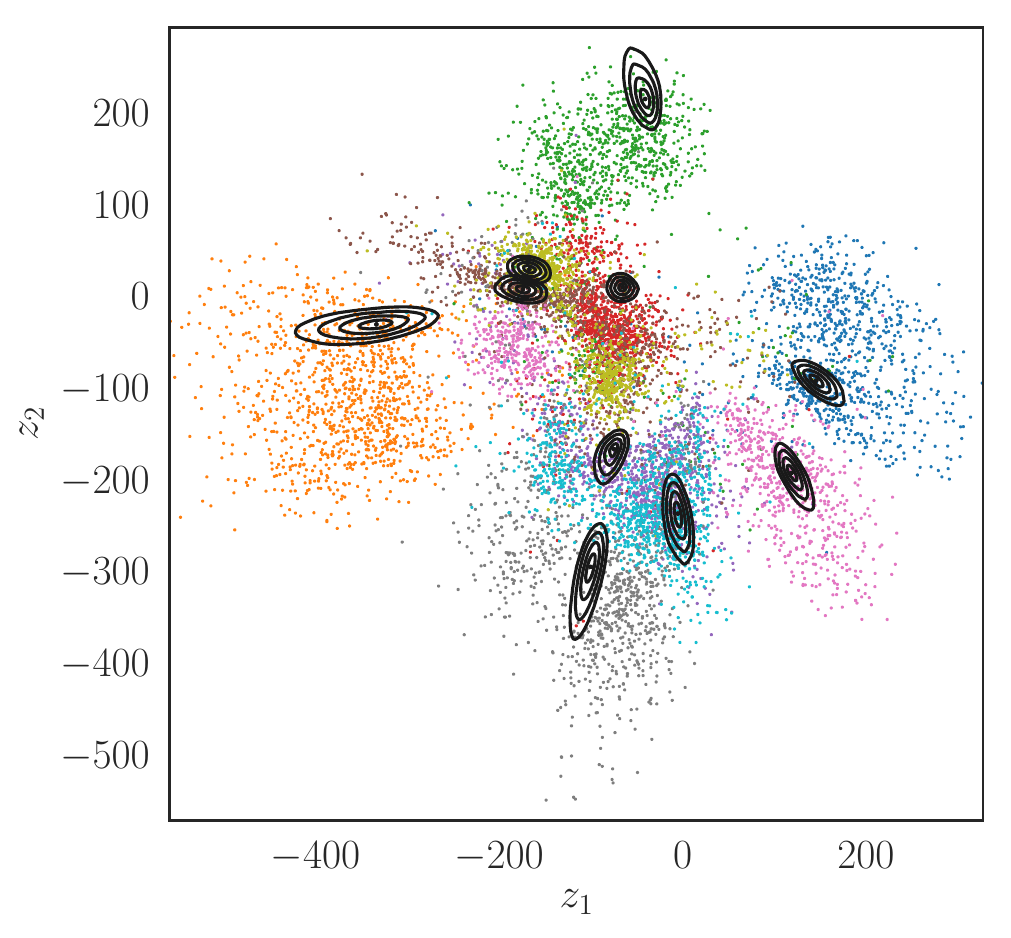}
        \caption{\methodreg}%
    \end{subfigure}
    \hfill
    \begin{subfigure}[b]{0.49\columnwidth}
        \includegraphics[width=\textwidth]{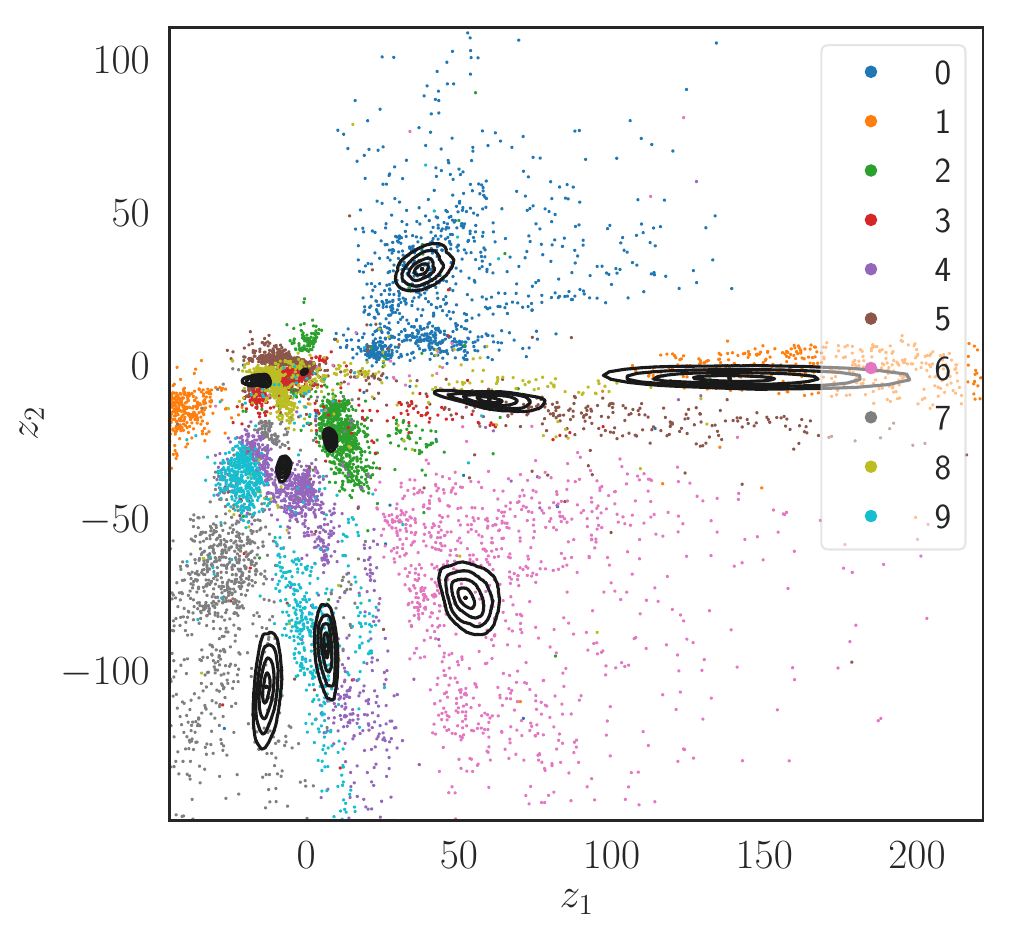}
        \caption{VHP-VAE}%
    \end{subfigure}
    \caption{Latent representation of MNIST data: the contour plots illustrate curves of equal \emph{observation-space distance} to the respective encoded data point (denoted by a black dot).}
    \label{fig:mnist_equal}
\end{figure}

\begin{figure}[b]
	\centering
	\includegraphics[width=0.12\textwidth]{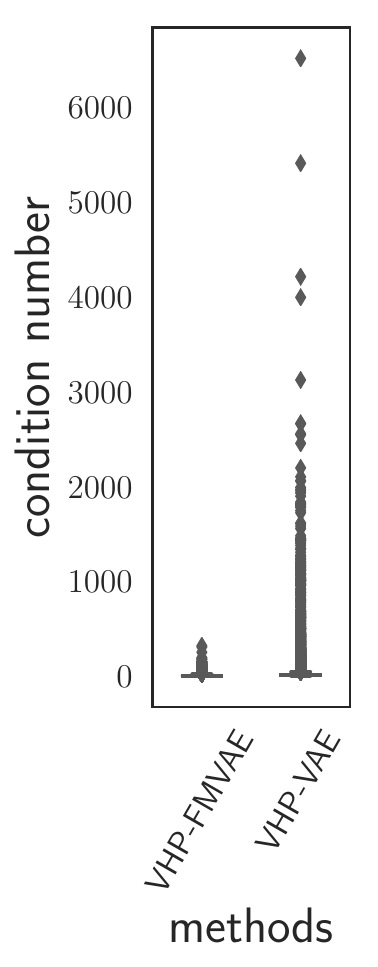}
	\includegraphics[width=0.12\textwidth]{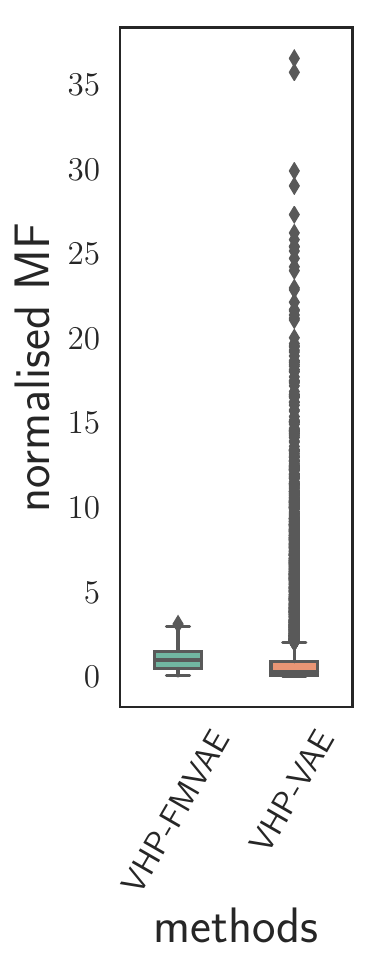}
	\caption{MNIST data: if both the condition number and the normalised MF values are close to one, it indicates that $\G(\z)\propto\mathds{1}$. The box-plots are based on 10,000 generated samples.}%
	\label{fig:mnist_cn_mf}
\end{figure}

Both of our evaluation metrics the condition number and the normalised $\MF$ show that the \methodreg outperforms the VAE-VHP (see Fig.~\ref{fig:mnist_equal} and Fig.~\ref{fig:mnist_cn_mf}).
In contrast to the VHP-VAE, the \methodreg learns a latent space, where Euclidean distances are close to geodesic distances (see Tab.~\ref{tab:geodesic}).
This indicates that $\G(\z)$ is approximately constant.

\subsection{MOT16 Object-Tracking Database}

We evaluate our approach on the MOT16 object-tracking database \citep{milan2016mot16}, which is a large-scale person re-identification data-set, containing both static and dynamic scenes from diverse cameras.

\setlength{\tabcolsep}{3.5pt}
\begin{table*}[ht!]
\vskip -0.1in
\caption{Comparisons between different descriptors for the purposes of object tracking and re-identification \citep{ristani2016performancemeasures}. The bold and the red numbers denote the best results among all methods and among non-supervised methods, respectively.}
\label{tab:sort}
\begin{center}
\begin{footnotesize}
\begin{sc}
\begin{tabular}{l|lccccccccccccccccc}
\toprule
Method                                                  					& Type        	 & IDF$_1$$\uparrow$ & IDP$\uparrow$  & IDR$\uparrow$ & Recall$\uparrow$  & Precision$\uparrow$    & FAR$\downarrow$   & MT$\uparrow$       \\
\midrule	
\begin{tabular}[c]{@{}l@{}}\methodreg-SORT $\eta=300$ (ours)\end{tabular}  & unsupervised & 63.7          	& 77.0          & 54.3          & 65.0          & 92.3          	& \textcolor{red}{\textbf{1.12}}  & 158    \\ 
\begin{tabular}[c]{@{}l@{}}\methodreg-SORT $\eta=3000$ (ours)\end{tabular} & unsupervised & \textcolor{red}{64.2}          	& \textcolor{red}{\textbf{77.6}} & \textcolor{red}{54.8}          & 65.1       & \textcolor{red}{\textbf{92.3}} & 1.13 & 162    \\ 
VHP-VAE-SORT                                                   				 & unsupervised & 60.5         	 & 72.3          & 52.1          & 65.8          & 91.4          	& 1.28 & \textcolor{red}{170}        \\ 
\midrule
SORT                                                        					& n.a.         	 &57.0         	 & 67.4          & 49.4          & \textcolor{red}{66.4}          & 90.6          & 1.44  & 158           \\ 
\midrule
DeepSORT                                                    				& supervised   	& \textbf{64.7} 	& 76.9          & \textbf{55.8} & \textbf{66.7} & 91.9     & 1.22  & \textbf{180}  \\ 
\bottomrule
\end{tabular}
\end{sc}
\vskip 0.1in
\begin{sc}
\begin{tabular}{l|lccccccccccccccccc}
\toprule
Method                                                  					  & PT$\downarrow$  	& ML$\downarrow$   & FP$\downarrow$ & FN$\downarrow$   & IDs$\downarrow$	& FM$\downarrow$   & MOTA $\uparrow$ & MOTP $\uparrow$ & MOTAL$\uparrow$       \\
\midrule	
\begin{tabular}[c]{@{}l@{}}\methodreg-SORT $\eta=300$ (ours)\end{tabular}  & 269          	& 90          & \textcolor{red}{\textbf{5950}} & 38592          & \textcolor{red}{616}          	& \textcolor{red}{\textbf{1143}} & 59.1       & 81.8 	& 59.7          \\ 
%\midrule
\begin{tabular}[c]{@{}l@{}}\methodreg-SORT $\eta=3000$ (ours)\end{tabular}& \textcolor{red}{265}          	& 90          & 6026          	& 38515          & 598          	& 1163          & \textcolor{red}{59.1}          & 81.8 	& \textcolor{red}{59.7}          \\
VHP-VAE-SORT                                                   				 & 266          	& \textcolor{red}{\textbf{81}} & 6820          & 37739          & 693          	& 1264          & 59.0          & 81.6          	& 59.6          \\  
\midrule
SORT                                                        					  & 275         	 & 84          & 7643          	& \textcolor{red}{37071}          & 1486         	& 1515          & 58.2          	& \textcolor{red}{\textbf{81.9}}          	& 59.5          \\ 
\midrule
DeepSORT                                                    				 & \textbf{250}	&87          & 6506          	& \textbf{36747} & \textbf{585} & 1165          & \textbf{60.3} & 81.6          	& \textbf{60.8} \\ 
%\midrule
\bottomrule
\end{tabular}
\end{sc}
\end{footnotesize}
\end{center}
\vskip -0.1in
\end{table*}

\begin{figure}[ht!]
 \centering
\begin{subfigure}{.42\textwidth}
  \centering
  \includegraphics[width=0.28\linewidth]{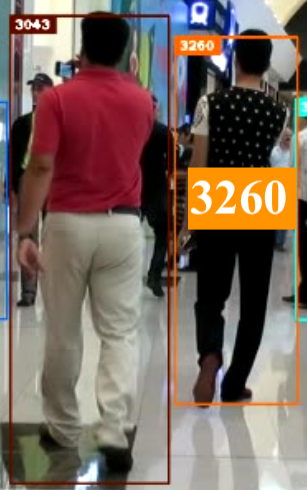}
  \includegraphics[width=0.208\linewidth]{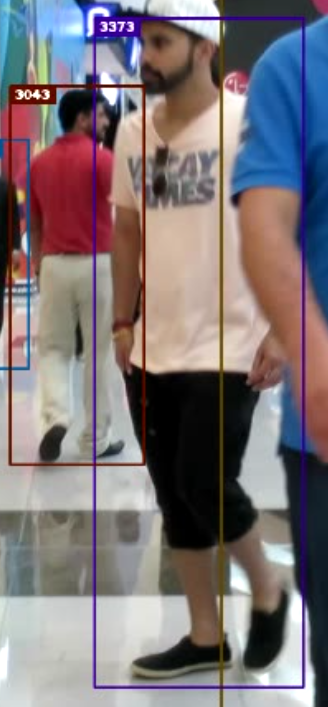}
  \includegraphics[width=0.317\linewidth]{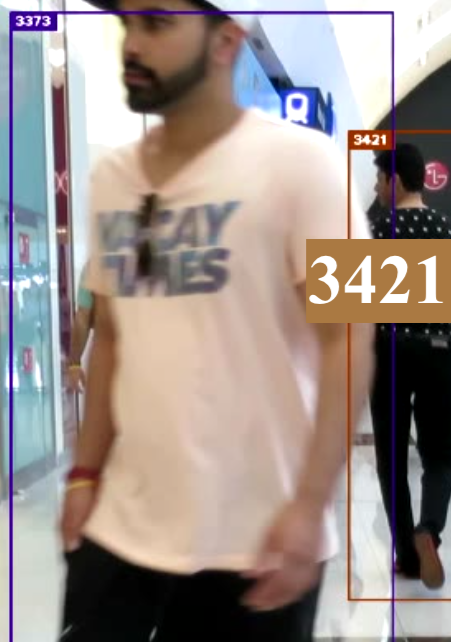}
  \caption{SORT}
  \label{fig:sort_3}
\end{subfigure}
\hfill
\begin{subfigure}{.42\textwidth}
  \centering
  \includegraphics[width=0.28\linewidth]{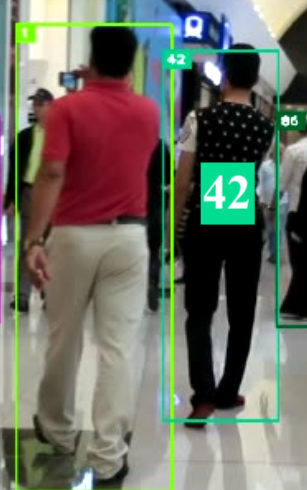}
  \includegraphics[width=0.208\linewidth]{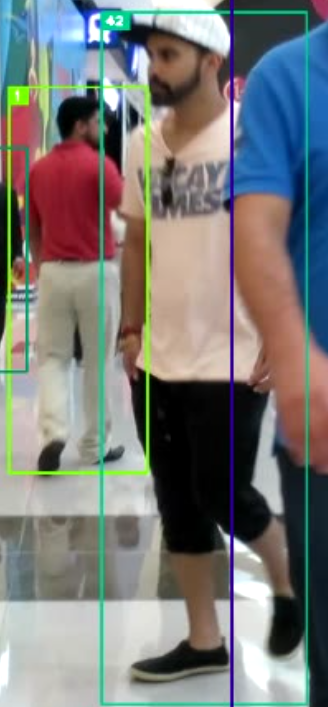}
  \includegraphics[width=0.317\linewidth]{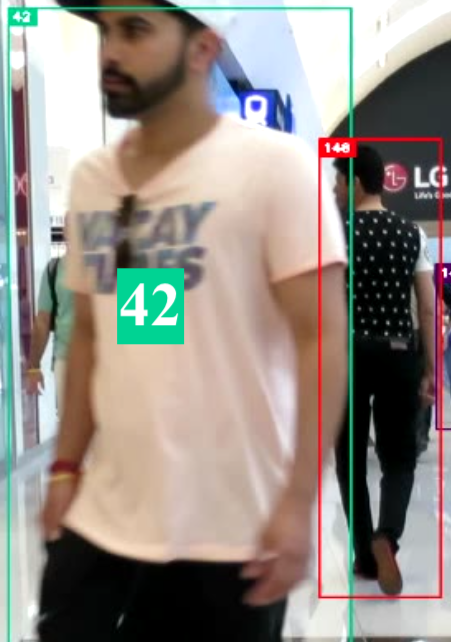}
  \caption{DeepSORT}
  \label{fig:deep_sort_3}
\end{subfigure}
\\
\begin{subfigure}{.42\textwidth}
  \centering
  \includegraphics[width=0.28\linewidth]{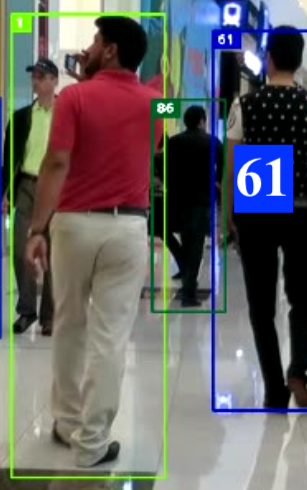}
  \includegraphics[width=0.208\linewidth]{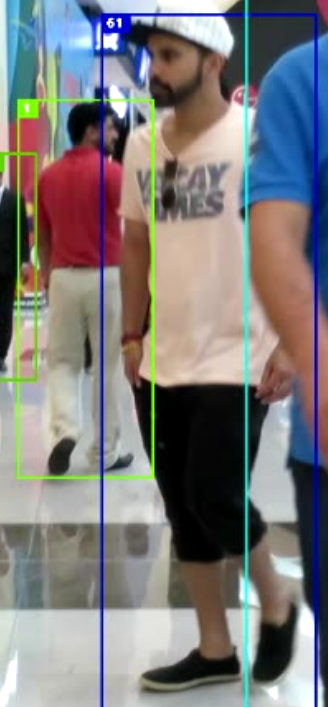}
  \includegraphics[width=0.317\linewidth]{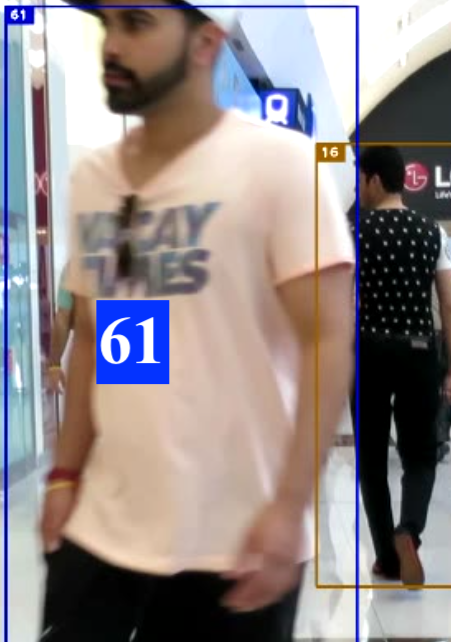}
  \caption{VHP-VAE-SORT}
  \label{fig:vhp_3}
\end{subfigure}
\hfill
\begin{subfigure}{.42\textwidth}
  \centering
  \includegraphics[width=0.28\linewidth]{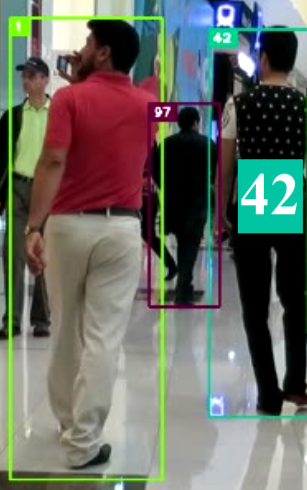}
   \includegraphics[width=0.208\linewidth]{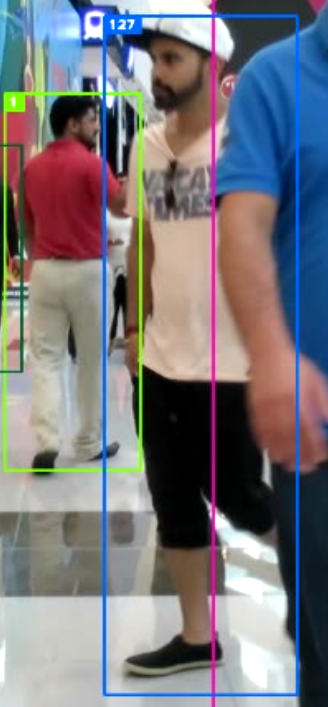}
  \includegraphics[width=0.317\linewidth]{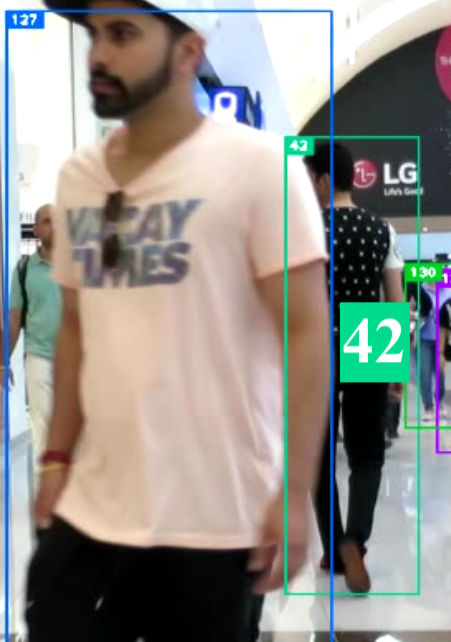}
  \caption{\methodreg-SORT with $\eta=3000$}
  \label{fig:vhp_reg_3}
\end{subfigure}
\caption{Example identity switches between overlapping tracks. For vanilla SORT, track 3260 gets occluded and when subsequently visible, it gets assigned a new ID 3421. For deeSORT and VHP-VAE-SORT, the occluding track gets assigned the same ID as the track it occludes (42/61), and subsequently keeps this (erroneous) track. For \methodreg-SORT, the track 42 gets occluded, but is re-identified correctly when again visible.} 
\label{fig:tracking_fig}
\end{figure}

We compare with two baselines: SORT \citep{bewley2016simple} and DeepSORT \citep{wojke2017simple}. SORT is a simple online and real-time tracking method, which uses bounding box intersection-over-union (IOU) for associating detections between frames and Kalman filters for the track predictions. It relies on good two-dimensional bounding box detections from a separate detector, and suffers from ID switching when tracks overlap in the image. DeepSORT extends the original SORT algorithm to integrate appearance information based on a deep appearance descriptor, which helps with re-identification in the case of such overlaps or missed detections. The deep appearance descriptor is trained using a \textit{supervised} cosine metric learning approach \citep{wojke2018deep}. The candidate object locations of the pre-generated detections for both SORT, DeepSORT and our method are taken from \citep{yu2016mot}. Further details regarding the implementation can be found in App.~\ref{app:mot16}.

We use the following metrics for evaluation. $\uparrow$ indicates that the higher the score is, the better the performance is. On the contrary, $\downarrow$ indicates that the lower the score is, the better the performance is. 
\begin{multicols}{2}
$\cdot$ IDF$_1$($\uparrow$): ID F$_1$ Score\\
$\cdot$ IDP($\uparrow$):  ID Precision\\
$\cdot$ IDR($\uparrow$):  ID Recall\\
$\cdot$ FAR($\downarrow$): False Alarm Ratio\\
$\cdot$ MT($\uparrow$): Mostly Tracked Trajectory\\
$\cdot$ PT($\downarrow$): Partially Tracked Trajectory\\
$\cdot$ ML($\downarrow$): Mostly Lost Trajectory\\
$\cdot$ FP($\downarrow$): False Positives\\
$\cdot$ FN($\downarrow$): False Negatives\\
$\cdot$ IDs($\downarrow$): Number of times an ID switches to a different previously tracked object\\
$\cdot$ FM($\downarrow$): Fragmentations\\
$\cdot$ MOTA($\uparrow$): Multi-object tracking accuracy\\
$\cdot$ MOTP($\uparrow$): Multi-object tracking precision\\
$\cdot$ MOTAL($\uparrow$): Log tracking accuracy
\end{multicols}

Tab.~\ref{tab:sort} shows that the performance of the proposed method is better than that of the model without Jacobian regularisation, and even close to the the performance of supervised learning.
All methods depend on the same underlying detector for object candidates, and identical Kalman filter parameters. Compared to baseline SORT which does not utilise any appearance information, DeepSORT has 2.54 times, VHP-VAE-SORT has 2.14 times, \methodreg-SORT ($\eta=300$) has 2.41 times and \methodreg-SORT ($\eta=3000$) has 2.48 times fewer ID switches. Whilst the supervised DeepSORT descriptor has the least, using unsupervised VAEs with flat decoders has only 2.2\% more switches, without the need for labels. Furthermore, by ensuring a quasi-Euclidean latent space, one can query nearest-neighbours efficiently via data-structures such as kDTrees. Fig.~\ref{fig:tracking_fig} shows an example of the results. In other examples of the videos, the \methodreg-SORT works similar as the DeepSORT.
Videos of the results can be downloaded at: \url{http://tiny.cc/0s71cz}

%!TEX root = ../FMVAEs.tex

\section{Conclusion}

In this paper, we have proposed a novel approach, which we call \emph{flat manifold} variational auto-encoder.
We have shown that this class of VAEs learns a latent representation, where the Euclidean metric is a proxy for the similarity between data points.
This is realised by interpreting the latent space as a Riemannian manifold and by combining a powerful empirical Bayes prior with a regularisation method that constrains the Riemannian metric tensor to be a scaled identity matrix.
Experiments on several datasets have shown the effectiveness of our proposed algorithm for measuring similarity.
In case of the MOT16 object-tracking database, the performance of our unsupervised method nears that of state-of-the-art supervised approaches.

\newpage
\pagebreak

\section*{Acknowledgements}
 We would like to thank Botond Cseke and Alexandros Paraschos for helpful discussions.

\bibliography{mybib.bib}
\bibliographystyle{icml2020}

\newpage
\pagebreak

\appendix
%!TEX root = ../FMVAEs.tex
\newpage
\cleardoublepage
\appendix
\onecolumn

\section{Appendix}

\subsection{Additional Results on the Human Motion Dataset}

\subsection{Influence of $c^2$}
\label{app:compc}
\begin{figure}[!ht]
	\centering
	\begin{subfigure}[b]{0.35\columnwidth}
		\includegraphics[width=\textwidth]{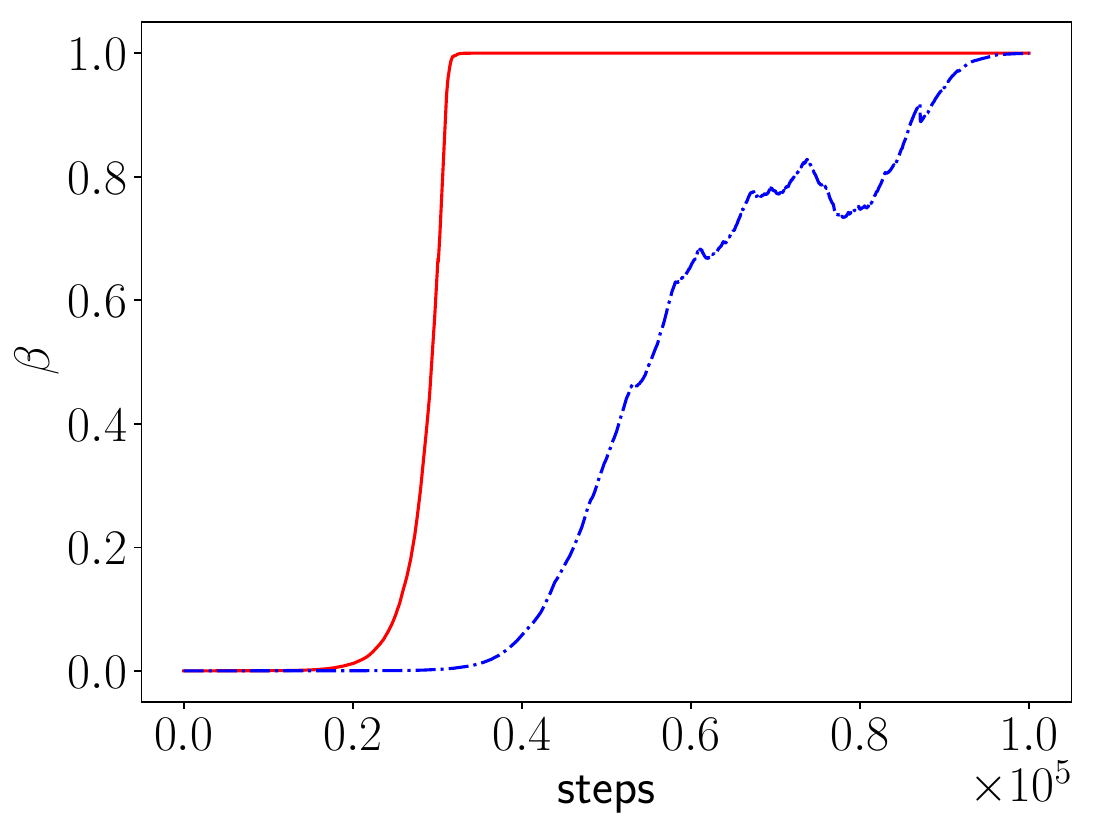}
		\caption{$\beta$}%
	\end{subfigure}
	%\hfill
	\begin{subfigure}[b]{0.35\columnwidth}
		\includegraphics[width=\textwidth]{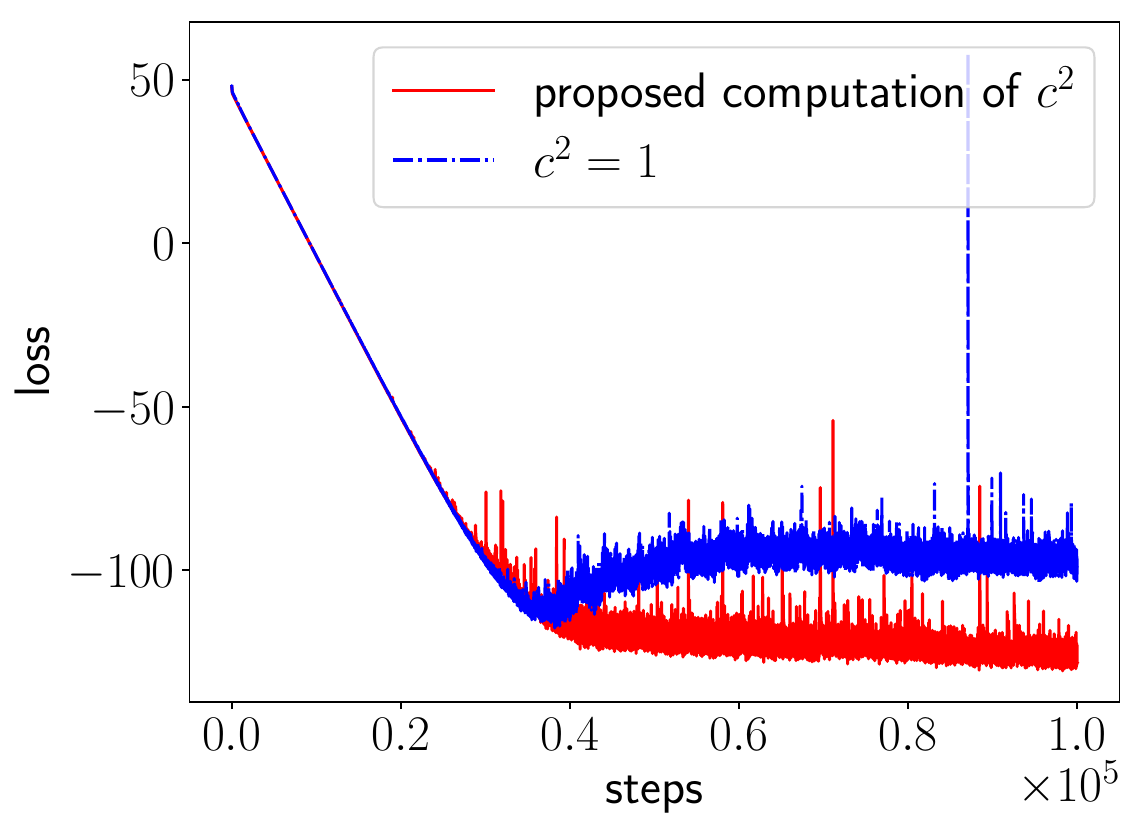}
		\caption{Losses of the training data set}%
	\end{subfigure}
	\caption{Comparison of different $c^2$ using the human motion dataset. The model with the proposed computation of $c^2$ converges faster than the model with $c^2 = 1$.}
\end{figure}

\subsubsection{Vector Field}
\begin{figure}[!ht]
    \centering
    \begin{subfigure}[b]{0.24\columnwidth}
        \includegraphics[width=\textwidth]{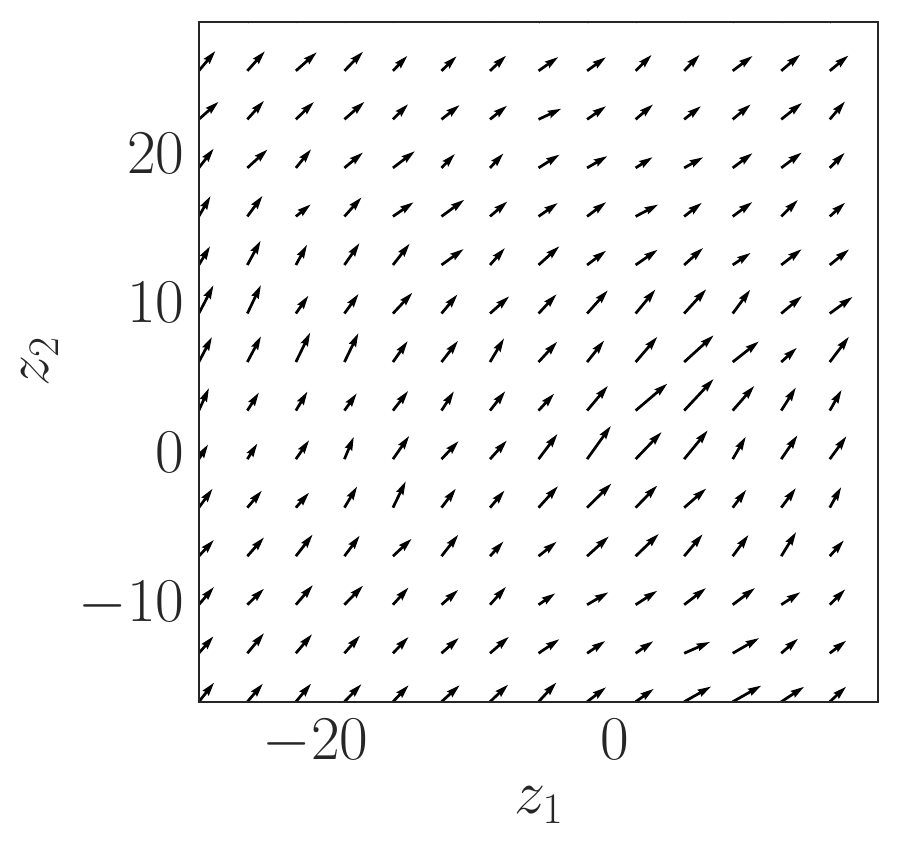}
        \caption{\methodreg.}%
    \end{subfigure}
    %\hfill
    \begin{subfigure}[b]{0.22\columnwidth}
        \includegraphics[width=\textwidth]{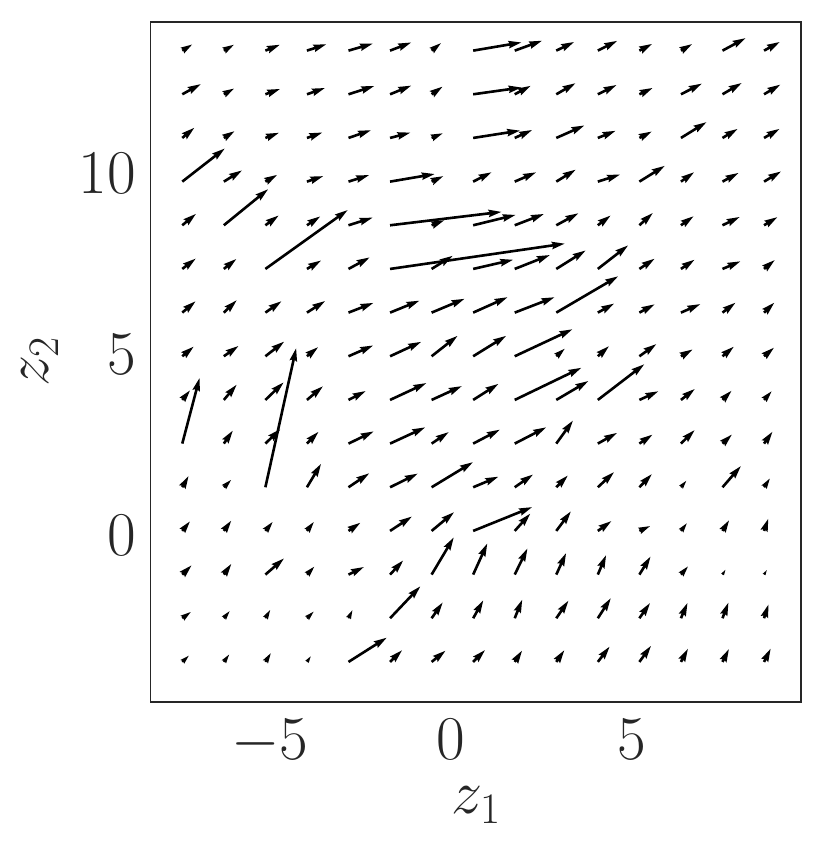}
        \caption{VHP-VAE.}%
    \end{subfigure}
    \caption{Vector field of the human motion dataset. The vector field is a vector of $L_2$ norm over the output of Jacobian. The figures are corresponding to Fig.~\ref{fig:human_equal}. The vector field of \methodreg is more regular than that of VAE-VHP.}
\end{figure}

%\subsubsection{Condition Number with 5D latent space}

%Human. motion m std 5d VHP 4.696235, 3.7197473

%Human. motion m std 5d proposed 2.5299563, 0.933556

\subsubsection{Results with a 5D Latent Space}
\label{5d_result}

For the comparison of the geodesic in Sec.~\ref{sec:human} (Tab.~\ref{tab:geodesic}) and App.~\ref{5d_result} (Tab.~\ref{table:5d}), the singular regularisation hyper-parameter (see the definition in Eq.~(17) of \cite{ChenKK2018metrics}), $\xi$, of the graph-based geodesic is scaled by $\xi_{\text{\methodreg}} = \frac{\text{mean}(s_{i\text{\methodreg}} \text{(data)})^2 \cdot \xi_{\text{VHP-VAE}}}{\text{mean}(s_{i\text{VHP-VAE}} \text{(data)})^2} $. $s_i$ denotes the singular of $\G$. $s_{i}(\text{data})$ is computed with training data. 
%The singular regularisation \cite{ChenKK2018metrics} is written as \begin{align}
%\hat{\G}=\mathbf{U}_r \mathrm{diag} \left\{ \frac{s_i^3}{s_i^2+\xi} \right\}^{r}_{i=1} \mathbf{V}_r^T.
%\end{align}

\begin{table}[ht!]
\vskip -0.1in
\caption{Verification of the distance metric with a 5D latent space. The table shows the length ratio of the Euclidean interpolation to the geodesic. Additionally, we list the ratio of the related distances in the observation space. Note: the ratio also depends on the hyper-parameter of the graph-based approach, $\xi$. Given a pair of \{$\xi_{\text{\methodreg}}$, $\xi_{\text{VHP-VAE}}$\} as computed in App.~\ref{5d_result}, the \methodreg outperforms the VHP-VAE.}
\begin{center}
\begin{footnotesize}
\begin{sc}
%\resizebox{0.75\textwidth}{!}{%					
\begin{tabular}{llcc}
\toprule
data-set	& 	method	& observation & latent \\
\midrule
Human 	& \methodreg	& \textbf{1.03 $\pm$ 0.16} 	& \textbf{0.59 $\pm$ 0.11} \\ 
		& VHP-VAE	& 1.36 $\pm$ 0.27 			& 0.47 $\pm$ 0.14\\
%\midrule
%MNIST	& \methodreg	&  \textbf{1.01 $\pm$ 0.08} 	&  \textbf{0.92 $\pm$ 0.05}\\
%		& VHP-VAE 	& 1.13 $\pm$0.22  			& 0.70 $\pm$ 0.31 \\
\bottomrule
\end{tabular}
%}
\end{sc}
\end{footnotesize}
\end{center}
\label{table:5d}
\vskip -0.1in
\end{table}

\begin{figure}
    \centering
    \begin{subfigure}[b]{0.15\columnwidth}
        \includegraphics[width=\textwidth]{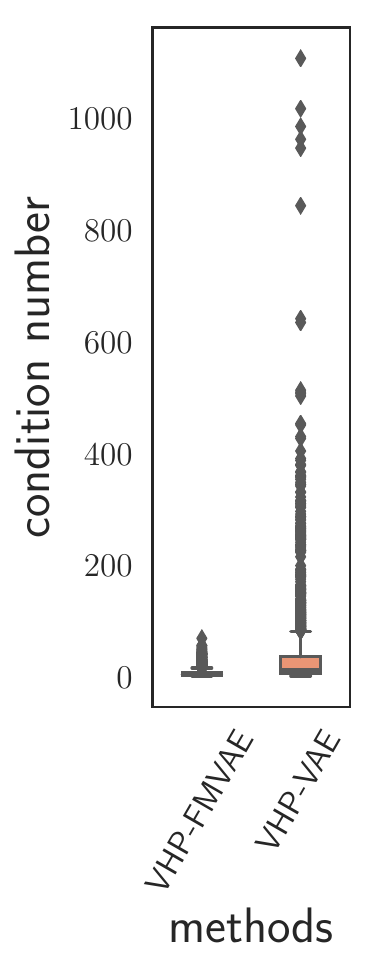}
        \label{fig:human_cn_5d}
    \end{subfigure}
    %\hfill
    \begin{subfigure}[b]{0.136\columnwidth}
        \includegraphics[width=\textwidth]{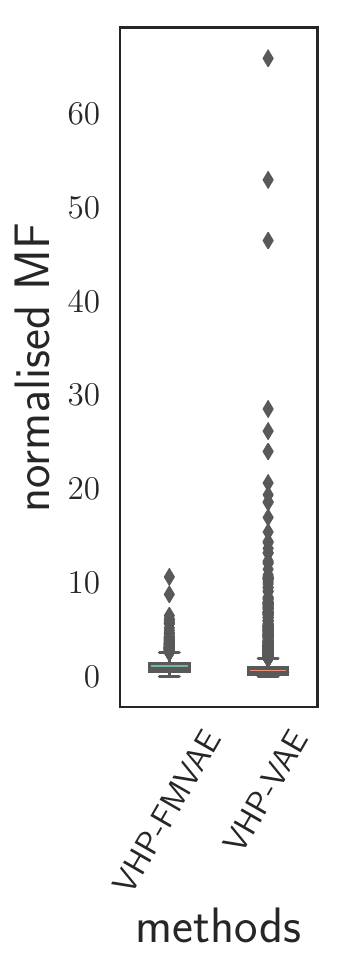}
        %\caption{Normalised MF.}%
        \label{fig:human_mf_5d}
    \end{subfigure}
	\vskip -0.1in
    \caption{Human motion data with a 5D latent spac: if both the condition number and the normalised MF values are close to one, it indicates that $\G(\z)\propto\mathds{1}$. The box-plots are based on 3,000 generated samples.}
\end{figure}

%\subsection{Additional Results on MNIST}

%\pagebreak
\subsection{Implementation of \methodreg-SORT}
\label{app:mot16}

SORT (Simple Object Real-time Tracking) uses 2D detections from a neural network and associates measurements of each frame to tracks that are initiated, kept, or removed over time. The IOU overlap is used as a distance function between a given track box and measurement box, and all boxes are optimally associated using the Hungarian algorithm.
DeepSORT is an extension of SORT wherein a ``deep'' association metric is added. This is learnt using a large person re-identification dataset, training a network that outputs a fixed vector output per object. This vector contains appearance information. During online application, the vector is used with nearest neighbor queries to establish measurement-to-track associations, instead of just the IOU overlap used by the vanilla SORT.
In our paper, we train variational auto-encoders and use the hidden latent space representation as a drop-in replacement to the fixed vector outputted by supervised network of DeepSORT, effectively only running the encoder during evaluation.

We evaluate the performance of our model by replacing the appearance descriptor from DeepSORT with the latent space embedding from the various auto-encoders used, using the same size of 128. The hyperparameters used were held constant: the minimum detection confidence of $0.3$, NMS max overlap of $0.7$, max cosine distance $0.2$, max appearance budget $100$. We tested a \methodreg, and our regularised \methodreg with $\eta=300$ and $\eta=3000$.

\newpage

\subsection{Optimisation Process}
\label{app:alg}
Note: to be in line with previous literature \citep[e.g.][]{higgins2017beta, sonderby2016}, we use the $\beta$-parametrisation of the Lagrange multiplier $\beta=\frac{1}{\lambda}$ in our experiments.

As introduced in \citep{vahiprior2019}, we apply the following $\beta$-update scheme:
\begin{align}
\label{eq:beta_update}
\beta_t = \beta_{t-1}\cdot\exp\big[\nu\cdot f_\beta(\beta_{t-1}, \hat{\text{C}}_t-\kappa^2;\tau) \cdot (\hat{\text{C}}_t-\kappa^2)\big],
\end{align}
where $f_\beta$ is defined as
\begin{align}
\label{eq:beta_update_restriction}
f_\beta(\beta, \delta; \tau) =
\big(1-H(\delta)\big)\cdot\tanh\big(\tau\cdot (\beta-1)\big) - H(\delta).
\end{align}
$H$ is the Heaviside function and $\tau$ a slope parameter.
\vspace{1in}

\begin{figure}[htb]
	\centering
	\begin{minipage}{.7\linewidth}
		%\centering
		\begin{algorithm}[H]
			\small
			\caption{\small \methodreg}
			\label{alg:flm_rewo}
			\begin{algorithmic}
				\STATE Initialise~ $t=1$
				\STATE Initialise~ $\beta\ll 1$
				\STATE Initialise~ \textsc{InitialPhase = True}
				\WHILE{training}
				\STATE Read current data batch $\mathbf{x}_\text{ba}$
				\STATE Sample from variational posterior $\mathbf{z}_\text{ba} \sim q_\phi(\mathbf{z}\vert\mathbf{x}_\text{ba})$
				\STATE Shuffle the samples from variational posterior $\mathbf{z}^\prime_\text{ba}=\mathrm{shuffle}(\mathbf{z}_\text{ba})$
				\STATE Augment data $\mathbf{z}^\text{aug}_\text{ba}=g(\mathbf{z}_\text{ba}, \mathbf{z}^\prime_\text{ba})$
				%	\vspace{0.03in}
				\STATE Compute  $c^2=\frac{1}{\text{batch\_size}}\sum_i\frac{1}{N_\z}\big[\mathrm{tr}(\G(\mathbf{z}^\text{aug}_i))\big]$
				%\STATE Compute $\mathop{\mathbb{E}_{\mathbf{x}_{i,j} \sim p_\mathcal{D}(\mathbf{x})}} \mathbb{E}_{\z_{i,j} \sim q_\phi(\mathbf{z}\vert\mathbf{x}_{i,j})}  \big[ \|  \G(g(\z_i, \z_j))  - c^2\mathds{1}\|_2^2 \big]$
				\STATE Compute $\hat{\text{C}}_\text{ba}$ (batch average)
				\STATE $\hat{\text{C}}_t = (1 -\alpha)\cdot\hat{\text{C}}_\text{ba} + \alpha\cdot\hat{\text{C}}_{t-1}$, ($\hat{\text{C}}_0 = \hat{\text{C}}_\text{ba}$)
				\IF{$\hat{\text{C}}_t < \kappa^2$}
				\STATE \textsc{InitialPhase = False}
				\ENDIF
				\IF{\textsc{InitialPhase}}
				\STATE Optimise $\mathcal{L}_\textsc{\methodreg}(\theta,\phi,\Theta,\Phi;\beta,\eta, c^2)$~~w.r.t~~~$\theta,\phi$	
				\ELSE
				\STATE $\beta \leftarrow
				\beta\cdot\exp\big[\nu\cdot f_\beta(\beta_{t-1}, \hat{\text{C}}_t-\kappa^2;\tau)\cdot(\hat{\text{C}}_t-\kappa^2)\big]$		
				\STATE Optimise $\mathcal{L}_\textsc{\methodreg}(\theta,\phi,\Theta,\Phi;\beta,\eta, c^2)$~~w.r.t~~~$\theta,\phi,\Theta,\Phi$		
				\ENDIF
				\STATE $t\leftarrow t+1$
				\ENDWHILE
			\end{algorithmic}
		\end{algorithm}
	\end{minipage}
\end{figure}
\vspace{1in}

\pagebreak
\subsection{Model Architectures}
\label{app:arch}
\begin{table}[!htb]
	\caption{Model architectures. FC refers to fully-connected layers. Conv2D and Conv2DT denote tow-D convolution layer and transposed two-D convolution layer, respectively. See the definition of $\nu$ in \citep{vahiprior2019}. We train each dataset on a single GPU.}
	\label{tab:hyp}
	\vskip 0.15in
	\begin{center}
		\begin{small}
			\begin{sc}
				\begin{tabular}{llll}
					\toprule
					Dataset		& 	Optimiser		& 	 Architecture		&			 \\
					\midrule
					Pendulum		& 	Adam		&	Input				&	16$\times$16$\times$1	 \\
					&	1$e$-4		&	Latents			&	2	\\
					&				&	$q_\phi(\mathbf{z}\vert\mathbf{x})$				& 	FC 256, 256. ReLU activation.		\\
					&				&	$p_\theta(\mathbf{x}\vert\mathbf{z})$				&	FC 256, 256. ReLU activation. Gaussian.		\\
					&				&	$q_\Phi(\mathbf{\zeta}\vert\mathbf{z})$				&	FC 256, 256, ReLU activation. 	\\
					&				&	$p_\Theta(\mathbf{z}\vert\mathbf{\zeta})$		&	FC 256, 256, ReLU activation. 	\\
					&				&	Others			& 	$\kappa$ = 0.025, $\nu$ = 1, $K$ = 16, $\eta = 1000$.\\
					\midrule
					CMU Human 		& 	Adam		&	Input				&	50		 \\
					&	1$e$-4		&	Latents			&	2, 5	\\
					&				&	$q_\phi(\mathbf{z}\vert\mathbf{x})$				& 	FC 256, 256, 256, 256. ReLU activation.		\\
					&				&	$p_\theta(\mathbf{x}\vert\mathbf{z})$				&	FC 256, 256, 256, 256. ReLU activation. Gaussian.		\\
					&				&	$q_\Phi(\mathbf{\zeta}\vert\mathbf{z})$				&	FC 256, 256, 256, 256, ReLU activation. 	\\
					&				&	$p_\Theta(\mathbf{z}\vert\mathbf{\zeta})$		&	FC 256, 256, 256, 256, ReLU activation. 	\\
					&				&	Others			& 	$\kappa$ = 0.03, $\nu$ = 1, $K$ = 32, $\eta = 8000$.\\
					\midrule
					MNIST 			&	Adam		&	Input				&	28$\times$28$\times$1 \\
					&	1$e$-4		&	Latents			&  2 \\
					&				&	$q_\phi(\mathbf{z}\vert\mathbf{x})$				&	FC 256, 256, 256, 256. ReLU activation. \\
					&				&	$p_\theta(\mathbf{x}\vert\mathbf{z})$				& FC 256, 256, 256, 256. ReLU activation. Bernoulli. \\
					&				&	$q_\Phi(\mathbf{\zeta}\vert\mathbf{z})$				&	FC 256, 256, 256, 256. ReLU activation. \\
					&				&	$p_\Theta(\mathbf{z}\vert\mathbf{\zeta})$		&	FC 256, 256, 256, 256. ReLU activation. \\
					&				&	others			&	$\kappa$ = 0.245 ,  $\nu$ = 1, $K$ = 16, $\eta = 8000$.\\
					\midrule
					MOT16   		&	Adam		&	Input				&	64$\times$64$\times$3 \\
					&	3$e$-5		&	Latents			& 	128\\
					&				&	$q_\phi(\mathbf{z}\vert\mathbf{x})$				&	 VGG16 \citep{zissermann2014vgg}\\
					&				&	$p_\theta(\mathbf{x}\vert\mathbf{z})$				& Conv2DT+Conv2D 256, 128, 64, 32, 16.  \\
					&				&											&ReLU activation. Gaussian.\\
					&				&	$q_\Phi(\mathbf{\zeta}\vert\mathbf{z})$				& FC 512, 512. ReLU activation.	 \\
					&				&	$p_\Theta(\mathbf{z}\vert\mathbf{\zeta})$		& FC 512, 512. ReLU activation.	\\
					&				&	others			&	$\kappa$ = 0.8 , $\nu$ = 1, $K$ = 8, $\eta = 300~\mathrm{or}~3000$.\\
					\midrule					
				\end{tabular}
			\end{sc}
		\end{small}
	\end{center}
	\vskip -0.1in
\end{table}

\end{document}